\documentclass{article}
\usepackage{iclr2027_conference,times}

\usepackage{amsmath,amsfonts,bm}

\def\eqref#1{equation~\ref{#1}}
\def\1{\bm{1}}

\DeclareMathAlphabet{\mathsfit}{\encodingdefault}{\sfdefault}{m}{sl}
\SetMathAlphabet{\mathsfit}{bold}{\encodingdefault}{\sfdefault}{bx}{n}

\usepackage{hyperref}
\usepackage{url}
\usepackage{booktabs}
\usepackage{amsfonts}
\usepackage{amsmath}
\usepackage{microtype}
\usepackage{xcolor}
\usepackage{graphicx}
\usepackage{enumitem}

\newcommand{\nicefrac}[2]{{#1}/{#2}}
\newcommand{\method}{adapter thicket}
\newcommand{\single}{GRPO-single-adapter}
\newcommand{\singlem}{GRPO-single (matched)}
\newcommand{\corr}{\bar{\rho}}
\newcommand{\eb}[1]{{\tiny\,$\pm$#1}}

 \iclrfinalcopy % Uncomment for camera-ready version, but NOT for submission.

\title{Adapter Thickets: Splitting an RLVR Budget\\Beats Concentrating It}

\author{
  Jonathan Williams \\
  Princeton University \\
  \And
  Esin Tureci \\
  Princeton University \\
  \And
   Karthik R. Narasimhan \\
  Princeton University \\
}

\begin{document}

\maketitle
\fancyhead{}                       % Remove the conference header
\renewcommand{\headrulewidth}{0pt}  % Remove the horizontal header line

\begin{abstract}
Majority voting over sampled completions is the workhorse of test-time scaling, and reinforcement learning with verifiable rewards (RLVR) is the workhorse for making each completion better. The standard pipeline composes the two: train one policy with RLVR, then sample it many times and vote. We show that this composition is lossy. A vote can only overturn mistakes that its voters do not share, and RLVR sharpens a policy so that its samples increasingly make the same mistakes. With every method drawing exactly $160$ completions per problem, training a single LoRA adapter on the full RLVR budget raises single-sample accuracy on every model we test ($1.5$B--$8$B). Yet on three of four models it leaves the majority vote below that of the untrained base model, by up to $4.8$ points. The damage builds during training: voter errors grow steadily more correlated, and the majority vote accuracy peaks early before falling by up to $7.0$ points. The cause is concentration, not RLVR itself. We split the same data and training budget across $K$ LoRA adapters, each trained on its own random disjoint shard, and call the result an \emph{\method{}}. Thickets out-vote the fully trained adapter in all $16$ (model, $K$) settings, and for $K{\geq}4$ they stay within $0.8$ points of the base model or above it. A single adapter stopped early, at a thicket member's step count, is a strong control that matches thickets for small $K$. For $K{\geq}8$, thickets keep more of RLVR's single-sample gain and out-vote this control in six of eight settings. The cost of concentration also grows with the number of votes: from $16$ to $160$ votes, the thicket's lead over the fully trained adapter widens from $1.3$ to $3.3$ points. When the plan is to sample and vote, an RLVR budget is better spent broad than deep.

\end{abstract}

\section{Introduction}
\label{sec:intro}

Test-time scaling has many sophisticated instantiations, but the workhorse remains the simplest one. We sample $N$ completions and return the answer that appears most often~\citep{wang2023selfconsistency}. This self-consistency majority vote needs no verifier and no reward model, and it is a surprisingly hard baseline to beat~\citep{counting-hard-to-beat, feng2025prm}. Training has a workhorse too. Reinforcement learning with verifiable rewards (RLVR)~\citep{shao2024deepseekmath,guo2025deepseekr1} reliably raises the accuracy of a single sample. The standard pipeline composes the two: train one policy with RLVR, then sample it many times and vote. This paper asks whether that composition works. Holding the inference budget fixed, does spending an RLVR budget on one policy improve its vote?

Ensemble theory suggests caution. A good voter pool needs members that are individually competent and collectively diverse. When one voter is wrong, the others should not be wrong in the same way~\citep{lakshminarayanan2017deepensembles}. A vote can only overturn a mistake that the majority does not share, so what majority voting really pays for is \emph{decorrelated errors}. RLVR works by sharpening the policy around behaviors that already earn reward. Pass@1 rises, but the sampling distribution narrows and diversity collapses~\citep{yue2025rlvrboundary,dang2025weight,wen2025cotpassk,li2025choice}. Resampling one sharpened policy therefore produces voters that make the same mistakes. Prior work measures this collapse through pass@$k$ and coverage. We measure what it does to the majority vote.

We find that concentrated RLVR hurts the vote, and that the fix is simple. The damage is not inherent to RLVR. It comes from concentrating the whole budget in one policy. We therefore split the training pool into $K$ disjoint shards and train $K$ independent LoRA adapters~\citep{hu2022lora}, each with $\nicefrac{1}{K}$ of the rollout budget. This uses the same total data and the same total training compute as the single adapter. We call the result an \emph{\method{}}, echoing the finding that diverse task specialists sit densely around pretrained weights~\citep{gan2026neural}. Four design choices keep the comparison honest. Every configuration pools exactly $160$ completions per problem ($K$ voters $\times$ $\nicefrac{160}{K}$ samples). A \emph{step-matched} control stops the same single-adapter run at the step count a thicket member receives. We trace the single adapter's vote along its whole training run. And we recount every vote over each method's first $B$ completions, for $B$ from $16$ to $160$. We report six findings.
\begin{enumerate}[leftmargin=1.35em,itemsep=0.1em,topsep=0.2em]
    \item \textbf{Concentrated RLVR hurts the vote.} A single adapter trained on the full RLVR budget raises single-sample accuracy on every model, yet votes below the untrained base on three of four models (by up to $4.8$ p.p.) and covers less of the answer space than the base in seven of eight model--suite settings (Section~\ref{sec:results-harm}).
    \item \textbf{The damage accrues with training.} Along the single adapter's run, voter error correlation $\corr$ rises monotonically on every model while voter accuracy rises overall. The vote peaks by step $125$ and then falls by $1.4$ to $7.0$ p.p. On Qwen2.5-3B and 7B, no checkpoint out-votes the base (Section~\ref{sec:results-dyn}).
    \item \textbf{Splitting the budget avoids the damage.} Thickets beat the fully trained adapter in all $16$ (model, $K$) majority-vote settings and all $32$ (model, suite, $K$) coverage settings. Once $K{\geq}4$, they stay within $0.8$ p.p.\ of the base or above it on every model. Against the step-matched control, they win $10$ of $16$ settings and tie one (Section~\ref{sec:results-split}).
    \item \textbf{\ldots\ while keeping most of the single-sample gain.} A single adapter stopped at a member's step count is a strong control at $K \leq 4$, winning four of eight vote comparisons. At $K \geq 8$ thickets win six of eight, tie one, and have more accurate voters on three of four models. Where RLVR adds real pass@1 (Qwen2.5-1.5B and Llama, $+7.3$ and $+9.7$ p.p.), thicket members keep $85$--$95\%$ of it at $K \leq 4$ and $63$--$78\%$ at $K \geq 8$, while the step-matched run only $8$--$66\%$ (Section~\ref{sec:results-keep}).
    \item \textbf{The cost of concentration grows with the vote budget.} Averaged over the $16$ settings, the fully trained adapter votes level with the base at $16$ votes ($+0.15$ p.p.) and $2.58$ p.p.\ below it at $160$. The thicket's lead over it grows from $+1.33$ to $+3.28$ p.p. The thicket's lead over the base shrinks, from $+1.48$ to $+0.71$ p.p.\ (Section~\ref{sec:budget}).
    \item \textbf{Random shards are enough.} Engineering the shards by subject does not help. At $K{=}7$, random disjoint shards vote at least as well as subject-specialist shards on all four models (Section~\ref{sec:subject}).
\end{enumerate}

\begin{figure}[t]
    \centering
    \includegraphics[width=0.9\linewidth]{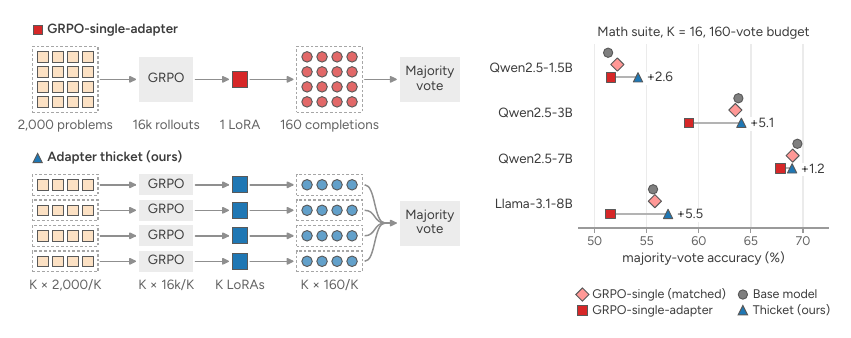}
    \caption{\textbf{One RLVR budget, one inference budget, two allocations.} \emph{Left:} \single{} trains one LoRA on the full $2{,}000$-problem pool with all $16$k rollouts. An adapter thicket trains $K$ LoRAs on disjoint $\nicefrac{2000}{K}$-example shards with $\nicefrac{16\mathrm{k}}{K}$ rollouts each. At inference every configuration pools exactly $160$ completions per problem ($K$ voters $\times \nicefrac{160}{K}$ samples). \emph{Right:} full-pool majority vote at $K{=}16$ on the math suite (macro mean over five benchmarks, three seeds), including the step-matched single-adapter control. The fully trained \single{} (red squares) falls below the base on three of four models. Deltas: thicket $-$ \single{}.}
    \label{fig:schematic}
\end{figure}

\section{Related work}
\label{sec:related}

Self-consistency majority voting~\citep{wang2023selfconsistency} remains the canonical verifier-free test-time scaling method. We treat it as the aggregator to feed rather than an opponent to replace.
Our results connect most directly to work showing that RLVR narrows the sampling distribution and reduces test-time coverage~\citep{yue2025rlvrboundary,li2025choice}. That work measures the collapse through pass@$k$. We measure its consequence for the majority vote at a fixed inference budget, and show that it can leave the vote below that of the untrained model. Prior approaches try to recover diversity within or around a single trained policy~\citep{dang2025weight}. We instead preserve it by splitting the training budget across independent policies.
Neural Thickets~\citep{gan2026neural} showed that full-weight
perturbations around a pretrained model can produce diverse specialists
that form an effective post-training ensemble. Existing LoRA ensemble
work has primarily targeted uncertainty estimation and
calibration~\citep{wang2023loraensembles,li2025elrea}, while related
methods compose, route, or merge already-trained
adapters~\citep{huang2023lorahub,wu2024mole,chronopoulou2023adaptersoup}.
We instead replace the perturb-and-select procedure of Neural Thickets
with reward-directed RLVR training in LoRA space and study how a fixed
RLVR budget should be allocated between one deeply trained adapter and
many shallowly trained adapters. We do not consider techniques that produce extra voters from a single run (e.g., snapshot ensembles~\citep{huang2017snapshot}) because they apply equally to both sides of our comparison.

\section{Adapter thickets and experimental setup}
\label{sec:method}

\textbf{Setup.} Let $\theta_0$ be a frozen instruction-tuned model and $\mathcal{P} = \{(x_j, a_j)\}_{j=1}^{D}$ a pool of problems with verifiable answers. A binary verifier $V(x, y, a) \in \{0,1\}$ checks the final answer of completion $y$ against $a$. All trained arms share one recipe: LoRA + GRPO with reward $r(x,y) = V(x,y,a)$, identical hyperparameters, and a total budget of $B_{\mathrm{train}}$ on-policy training rollouts. We compare four ways of fielding $K$ voters.
\begin{itemize}[leftmargin=1.35em,itemsep=0.1em,topsep=0.2em]
    \item \textbf{Base}: the untrained $\theta_0$, $K$ i.i.d.\ streams.
    \item \textbf{\single{}}: the standard recipe. One adapter trained on all of $\mathcal{P}$ with the full budget $B_{\mathrm{train}}$, sampled as $K$ i.i.d.\ streams from the final checkpoint.
    \item \textbf{\singlem{}}: the \emph{same} single-adapter run, evaluated at the checkpoint matched to a thicket member's step count (Appendix~\ref{app:hyperparams}), again sampled as $K$ i.i.d.\ streams. This control tests whether the thicket still wins when no voter is trained longer than a thicket member.
    \item \textbf{\method{}}: $K$ independently trained adapters, each trained for $\nicefrac{B_{\mathrm{train}}}{K}$ rollouts on one of $K$ disjoint random shards of $\mathcal{P}$. Each adapter is one voter.
\end{itemize}
Every training problem is seen exactly once, by one adapter, in both training arms.

\textbf{A fixed inference budget.} Each voter contributes $\nicefrac{160}{K}$ completions per problem, so \emph{every} method at \emph{every} $K$ pools exactly $160$ votes. Any difference between methods is therefore a difference in \emph{who votes}, never in how much is sampled. For a single policy, the $K$ voters are nominal. We draw one pool of $160$ i.i.d.\ completions per problem, each with its own sampling seed, and group it into $K$ streams. The base and \single{} votes are therefore identical at every $K$, and only the thicket and \singlem{} (whose checkpoint depends on $K$) change with $K$. Section~\ref{sec:budget} varies the budget from $16$ to $160$ votes. Before voting, we drop unparseable completions (nulls), merge mathematically equivalent answers (e.g., $0.5$ and $\nicefrac{1}{2}$) into one candidate, and marginalize exact ties (Appendix~\ref{app:eval}). We report four quantities: \textbf{majority vote} on the math suite, \textbf{coverage}~\citep{brown2024monkeys} (pass@pool, whether any of the $160$ completions is correct), \textbf{pass@1} (mean single-completion accuracy, which we also call voter accuracy), and voter diversity. Diversity is the mean pairwise Pearson correlation of the voters' binary error indicators, $\corr = \frac{1}{K(K-1)} \sum_{i \neq j} \mathrm{corr}(e_i, e_j)$.

\textbf{Seeds.} We repeat every main experiment under three seeds. The seed controls the thicket's data partition, the single adapter's data ordering, and the sampling seeds of every pool. Every number is a mean over the three seeds, with the sample standard deviation after $\pm$. The one exception is the subject-versus-random shard ablation (Section~\ref{sec:subject}), which uses seed $42$. Deltas quoted in prose are differences of the displayed means. A mean margin is the mean of these differences over settings.

\textbf{Training and evaluation.} We use Qwen2.5-\{1.5B, 3B, 7B\}-Instruct~\citep{yang2024qwen25} and Llama-3.1-8B-Instruct~\citep{grattafiori2024llama3}. The Qwen2.5 family spans the size axis. We add Llama-3.1-8B to confirm that the results generalize across model families rather than being a Qwen-specific artifact. We study two domains: \emph{math} ($D = 2{,}000$ MATH problems~\citep{hendrycks2021math}, with a verified-answer reward) and \emph{code} ($D = 2{,}000$ LeetCodeDataset problems~\citep{xia2025leetcodedataset}, with a unit-test reward). All runs use rank-$64$ LoRA, GRPO with $8$ rollouts per prompt ($B_{\mathrm{train}} = 16$k rollouts per arm), and $K \in \{2,4,8,16\}$ (Appendix~\ref{app:hyperparams}). Evaluation is zero-shot at temperature $1.0$ and top-$p$ $0.95$. The \emph{math} suite is MATH500, GSM8K~\citep{cobbe2021gsm8k}, OlympiadBench~\citep{he2024olympiadbench}, Countdown~\citep{katz2025countdown}, and Knights~\&~Knaves~\citep{xie2024knights}. The \emph{code} suite is MBPP~\citep{austin2021program}, HumanEval~\citep{chen2021codex}, and the LeetCodeDataset test split. Suite results are \emph{macro means over benchmarks}. Programs have no canonicalizable final answer to vote over, so code enters only the coverage, pass@1, and error-correlation analyses.

\section{Results}
\label{sec:results}

\begin{table}[t]
  \caption{\textbf{Majority-vote accuracy on the math suite under the fixed $160$-vote budget} (canonical-class votes; macro means (\%) over five benchmarks; mean\,$\pm$\,std over three seeds). Base and \single{} each vote over one $160$-completion pool, so their rows are constant across $K$ by construction. Bold: best method per model and $K$ by displayed mean; underlined: second best.}
  \label{tab:mv}
  \begin{center}
  \small
  \setlength{\tabcolsep}{5pt}
  \begin{tabular}{llcccc}
\toprule
Model & Method & $K{=}2$ & $K{=}4$ & $K{=}8$ & $K{=}16$ \\
\midrule
Qwen2.5-1.5B & Base & 51.3\eb{0.4} & 51.3\eb{0.4} & 51.3\eb{0.4} & 51.3\eb{0.4} \\
 & \singlem{} & \underline{54.6\eb{0.7}} & \textbf{55.3\eb{2.0}} & \underline{53.7\eb{0.2}} & \underline{52.2\eb{0.9}} \\
 & \single{} & 51.6\eb{1.6} & 51.6\eb{1.6} & 51.6\eb{1.6} & 51.6\eb{1.6} \\
 & Thicket & \textbf{55.2\eb{2.0}} & \underline{53.6\eb{0.7}} & \textbf{54.6\eb{0.5}} & \textbf{54.2\eb{0.2}} \\
\addlinespace[2pt]
Qwen2.5-3B & Base & \textbf{63.8\eb{0.3}} & \underline{63.8\eb{0.3}} & \textbf{63.8\eb{0.3}} & \underline{63.8\eb{0.3}} \\
 & \singlem{} & 61.9\eb{1.2} & 63.5\eb{1.2} & 63.3\eb{0.1} & 63.5\eb{0.8} \\
 & \single{} & 59.0\eb{5.1} & 59.0\eb{5.1} & 59.0\eb{5.1} & 59.0\eb{5.1} \\
 & Thicket & \underline{63.0\eb{1.2}} & \textbf{64.0\eb{0.4}} & \underline{63.6\eb{0.2}} & \textbf{64.1\eb{0.4}} \\
\addlinespace[2pt]
Qwen2.5-7B & Base & \textbf{69.5\eb{0.2}} & \textbf{69.5\eb{0.2}} & \textbf{69.5\eb{0.2}} & \textbf{69.5\eb{0.2}} \\
 & \singlem{} & \underline{68.7\eb{1.1}} & 67.9\eb{1.2} & \underline{69.2\eb{0.1}} & \underline{69.0\eb{0.2}} \\
 & \single{} & 67.8\eb{1.2} & 67.8\eb{1.2} & 67.8\eb{1.2} & 67.8\eb{1.2} \\
 & Thicket & 68.6\eb{1.1} & \underline{68.9\eb{0.4}} & 68.7\eb{1.0} & \underline{69.0\eb{0.4}} \\
\addlinespace[2pt]
Llama-3.1-8B & Base & \textbf{55.6\eb{0.3}} & 55.6\eb{0.3} & 55.6\eb{0.3} & 55.6\eb{0.3} \\
 & \singlem{} & \underline{53.9\eb{1.0}} & \textbf{58.5\eb{1.5}} & \underline{56.8\eb{0.5}} & \underline{55.8\eb{0.6}} \\
 & \single{} & 51.5\eb{3.2} & 51.5\eb{3.2} & 51.5\eb{3.2} & 51.5\eb{3.2} \\
 & Thicket & 53.0\eb{4.2} & \underline{56.8\eb{2.1}} & \textbf{57.8\eb{1.5}} & \textbf{57.0\eb{0.7}} \\
\bottomrule
\end{tabular}

  \end{center}
\end{table}

\subsection{Concentrated RLVR hurts the vote}
\label{sec:results-harm}

Table~\ref{tab:mv} gives the central comparison. Read the table in blocks of four rows. For a given model and voter count $K$, the four rows share one $160$-completion budget and differ only in who the voters are: fresh samples of the base model, samples of the single adapter stopped early (\singlem{}), samples of the fully trained single adapter (\single{}), or one sample stream from each of $K$ thicket members.

Start with the standard recipe. Full single-adapter RLVR does what it is supposed to do for a single sample: it raises pass@1 on every model, by $+7.3$, $+1.0$, $+0.7$, and $+9.7$ p.p.\ on Qwen2.5-1.5B, Qwen2.5-3B, Qwen2.5-7B, and Llama-3.1-8B. Yet at the same $160$-completion budget it votes \emph{below the untrained base} on Qwen2.5-3B ($-4.8$ p.p.), Qwen2.5-7B ($-1.7$), and Llama ($-4.1$), and only matches it on Qwen2.5-1.5B ($+0.3$). As work on diversity collapse would predict~\citep{yue2025rlvrboundary}, the lost votes track lost diversity. Full training raises the voters' error correlation $\corr$ above the base in all eight model--suite settings at $K{=}16$, by $+0.05$ to $+0.20$ (Table~\ref{tab:corr16} in Appendix~\ref{app:corr}). The fully trained adapter also covers \emph{less} of the answer space than the base in seven of eight model--suite settings, by up to $6.4$ p.p.\ on code, and ties it in the eighth (Table~\ref{tab:cov}). This deficit grows with the number of votes (Section~\ref{sec:budget}).

\subsection{The damage accrues with training}
\label{sec:results-dyn}

\begin{figure}[t]
    \centering
    \includegraphics[width=0.8\linewidth]{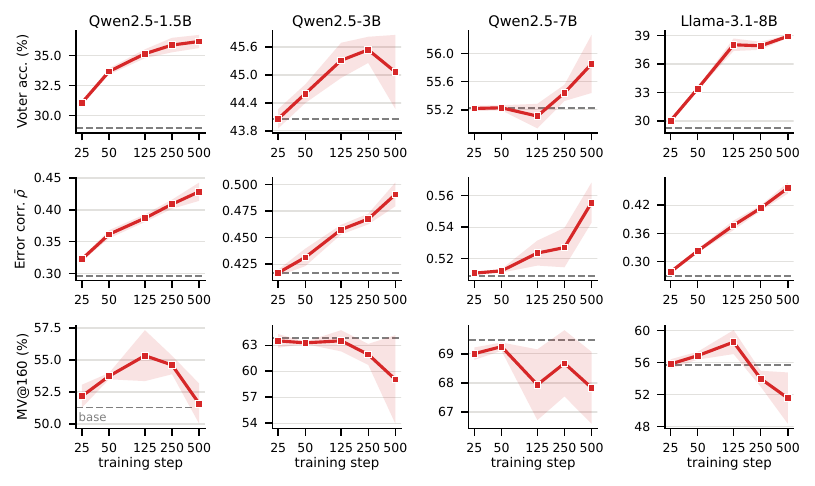}
    \caption{\textbf{Single-adapter training raises voter accuracy and voter correlation together -- and the vote pays for it.} Checkpoints of the \single{} run at steps $25/50/125/250$ and the final checkpoint (step $499$ for Qwen, $500$ for Llama; plotted at $500$), evaluated on the math suite with the same fixed $160$-completion budget per problem (grouped as $16$ voters $\times$ $10$ samples; mean over three seeds, $\pm 1$ std band). Top: mean voter accuracy. Middle: mean pairwise error correlation $\corr$. Bottom: majority vote over the $160$-vote pool. Dashed lines: the untrained base model. On every model $\corr$ rises monotonically with training, and the vote peaks by step $125$ before declining.}
    \label{fig:dynamics}
\end{figure}

Figure~\ref{fig:dynamics} traces the \single{} run along its training trajectory. We evaluate checkpoints at steps $25/50/125/250$ and the final checkpoint ($\approx 500$) under the same fixed $160$-vote budget. Mean voter accuracy rises overall, for example $30.0 \to 38.9$ on Llama and $31.1 \to 36.2$ on Qwen2.5-1.5B. But the voters' error correlation rises with it, monotonically on all four models: $0.28 \to 0.46$ on Llama and $0.32 \to 0.43$ on Qwen2.5-1.5B. The majority vote is what a test-time-scaling user actually consumes, and it peaks by step $125$ and then falls. Llama reaches $58.5$ at step $125$ and ends at $51.5$, a drop of $7.0$ p.p. Qwen2.5-1.5B peaks at $55.3$ (step $125$) and ends at $51.6$. On Qwen2.5-3B and 7B the vote never rises above the untrained base at all (peaks of $63.5$ and $69.2$, against base votes of $63.8$ and $69.5$), and training ends $4.8$ and $1.7$ p.p.\ below it.

This reframes the usual early-stopping question. If you plan to sample many times and vote, pass@1 is the wrong dial to watch, because the vote can be deteriorating while pass@1 still climbs. Stopping early can rescue the vote, but only at the right step, and the right step differs across models: step $125$ on Qwen2.5-1.5B and Llama, while on Qwen2.5-3B and 7B no checkpoint beats the base. Stopping early also leaves much of the training data and budget unused (three quarters of it at step $125$). The next section asks whether the budget can be spent in full without paying for it in the vote.

\subsection{Splitting the budget avoids the damage}
\label{sec:results-split}

\begin{table}[t]
  \caption{\textbf{Coverage} (pass@pool: at least one correct completion among the $160$) on the math and code suites; macro means (\%); mean\,$\pm$\,std over three seeds. Bold: best method per model, suite, and $K$; underlined: second best.}
  \label{tab:cov}
  \begin{center}
  \scriptsize
  \setlength{\tabcolsep}{2.6pt}
  \begin{tabular}{llcccccccc}
\toprule
 & & \multicolumn{4}{c}{Coverage, math} & \multicolumn{4}{c}{Coverage, code} \\
\cmidrule(lr){3-6} \cmidrule(lr){7-10}
Model & Method ($K{=}$) & 2 & 4 & 8 & 16 & 2 & 4 & 8 & 16 \\
\midrule
Qwen2.5-1.5B & Base & \textbf{89.6\eb{0.3}} & \textbf{89.6\eb{0.3}} & \underline{89.6\eb{0.3}} & \underline{89.6\eb{0.3}} & \textbf{72.2\eb{0.4}} & \textbf{72.2\eb{0.4}} & \textbf{72.2\eb{0.4}} & \textbf{72.2\eb{0.4}} \\
 & \singlem{} & \textbf{89.6\eb{0.4}} & 89.3\eb{0.5} & \textbf{90.0\eb{0.3}} & 89.5\eb{0.5} & 69.0\eb{1.8} & 70.2\eb{0.9} & \underline{72.0\eb{0.5}} & \underline{71.9\eb{0.3}} \\
 & \single{} & 87.7\eb{0.7} & 87.7\eb{0.7} & 87.7\eb{0.7} & 87.7\eb{0.7} & 65.8\eb{1.4} & 65.8\eb{1.4} & 65.8\eb{1.4} & 65.8\eb{1.4} \\
 & Thicket & \underline{88.8\eb{1.0}} & \underline{89.5\eb{0.4}} & \underline{89.6\eb{0.5}} & \textbf{89.8\eb{0.6}} & \underline{69.2\eb{1.4}} & \underline{70.6\eb{0.6}} & 70.6\eb{0.5} & 70.3\eb{0.3} \\
\addlinespace[2pt]
Qwen2.5-3B & Base & \textbf{91.7\eb{0.1}} & \textbf{91.7\eb{0.1}} & \textbf{91.7\eb{0.1}} & \textbf{91.7\eb{0.1}} & \textbf{75.6\eb{1.1}} & \textbf{75.6\eb{1.1}} & \textbf{75.6\eb{1.1}} & 75.6\eb{1.1} \\
 & \singlem{} & \underline{90.8\eb{0.3}} & \underline{91.0\eb{0.5}} & \underline{91.6\eb{0.6}} & \underline{91.6\eb{0.5}} & \underline{73.7\eb{2.0}} & 73.1\eb{1.7} & \underline{75.4\eb{0.8}} & \textbf{76.4\eb{0.9}} \\
 & \single{} & 88.9\eb{1.0} & 88.9\eb{1.0} & 88.9\eb{1.0} & 88.9\eb{1.0} & 70.7\eb{0.5} & 70.7\eb{0.5} & 70.7\eb{0.5} & 70.7\eb{0.5} \\
 & Thicket & \underline{90.8\eb{0.6}} & 90.8\eb{0.8} & 91.3\eb{0.4} & \underline{91.6\eb{0.1}} & 73.5\eb{1.4} & \underline{74.0\eb{1.4}} & 75.2\eb{0.4} & \underline{75.9\eb{0.3}} \\
\addlinespace[2pt]
Qwen2.5-7B & Base & \textbf{92.8\eb{0.4}} & \textbf{92.8\eb{0.4}} & \textbf{92.8\eb{0.4}} & \textbf{92.8\eb{0.4}} & \textbf{76.3\eb{0.2}} & \textbf{76.3\eb{0.2}} & \textbf{76.3\eb{0.2}} & \underline{76.3\eb{0.2}} \\
 & \singlem{} & \underline{92.3\eb{0.2}} & 91.7\eb{0.6} & \underline{92.6\eb{0.2}} & 92.3\eb{0.8} & 74.3\eb{0.5} & 75.7\eb{0.7} & \textbf{76.3\eb{0.4}} & \textbf{76.4\eb{0.2}} \\
 & \single{} & 91.2\eb{0.5} & 91.2\eb{0.5} & 91.2\eb{0.5} & 91.2\eb{0.5} & 73.1\eb{0.4} & 73.1\eb{0.4} & \underline{73.1\eb{0.4}} & 73.1\eb{0.4} \\
 & Thicket & 92.0\eb{0.2} & \underline{92.4\eb{0.4}} & 92.3\eb{0.2} & \underline{92.6\eb{0.2}} & \underline{75.7\eb{0.2}} & \underline{75.9\eb{0.2}} & \textbf{76.3\eb{0.4}} & \textbf{76.4\eb{0.4}} \\
\addlinespace[2pt]
Llama-3.1-8B & Base & 86.3\eb{0.3} & 86.3\eb{0.3} & 86.3\eb{0.3} & 86.3\eb{0.3} & \textbf{75.2\eb{0.2}} & \textbf{75.2\eb{0.2}} & \textbf{75.2\eb{0.2}} & \underline{75.2\eb{0.2}} \\
 & \singlem{} & \underline{87.5\eb{0.3}} & \textbf{89.4\eb{0.5}} & \underline{87.9\eb{0.1}} & \underline{86.8\eb{0.6}} & 72.6\eb{1.7} & 73.8\eb{1.1} & \underline{75.0\eb{0.4}} & \textbf{75.4\eb{0.1}} \\
 & \single{} & 86.3\eb{2.0} & 86.3\eb{2.0} & 86.3\eb{2.0} & 86.3\eb{2.0} & 71.8\eb{1.9} & 71.8\eb{1.9} & 71.8\eb{1.9} & 71.8\eb{1.9} \\
 & Thicket & \textbf{88.2\eb{0.2}} & \underline{89.1\eb{0.5}} & \textbf{89.9\eb{0.7}} & \textbf{88.5\eb{0.2}} & \underline{73.3\eb{1.4}} & \underline{74.4\eb{0.5}} & 74.3\eb{0.1} & 74.8\eb{0.5} \\
\bottomrule
\end{tabular}

  \end{center}
\end{table}

A thicket spends the same budget as \single{} on the same data, but spreads it across $K$ members. At the fixed $160$-vote budget, thickets beat the fully trained \single{} in \emph{all sixteen} (model, $K$) settings (Table~\ref{tab:mv}). The margins range from $+0.8$ to $+6.3$ p.p. At $K{=}16$ they are $+2.6$, $+5.1$, $+1.2$, and $+5.5$ p.p.\ on Qwen2.5-1.5B, Qwen2.5-3B, Qwen2.5-7B, and Llama. Coverage follows the vote. Thickets cover more than \single{} in all $32$ (model, suite, $K$) settings, by $+0.8$ to $+3.6$ p.p.\ on math and $+1.5$ to $+5.2$ p.p.\ on code (Table~\ref{tab:cov}).

Against the untrained base, the thicket's job is to avoid the damage, and it mostly does. Thickets beat the base in $9$ of $16$ settings, including every $K$ on Qwen2.5-1.5B (by $+2.3$ to $+3.9$) and every $K{\geq}4$ on Llama ($+1.2$ to $+2.2$). On Qwen2.5-3B they stay within $0.8$ p.p.\ of it, and on Qwen2.5-7B they trail it at every $K$, by $0.5$ to $0.9$ p.p. The one setting where a thicket trails the base by more than a point is the two-member Llama thicket ($53.0$ vs.\ $55.6$). A two-member committee is the least diverse thicket, since each member casts half the votes, and this cell has the largest seed spread of any thicket entry in Table~\ref{tab:mv} ($\pm 4.2$). Coverage tells the same story. Thickets stay close to base-level coverage, though usually just below it ($23$ of $32$ settings, by at most $3.0$ p.p.), and on Llama math they exceed it at every $K$.

The step-matched control is the stronger opponent. Thickets beat \singlem{} in $10$ of $16$ settings and tie one. Its two largest wins ($1.7$ p.p.\ each) come at $K{=}4$ on Qwen2.5-1.5B and Llama, and its other wins are at most $0.9$ p.p.\ (Section~\ref{sec:results-keep}). Per-benchmark exceptions to the thicket's win over \single{} are rare: $6$ of $80$ cells, all on Countdown or Knights~\&~Knaves.

\subsection{\ldots\ while keeping most of the single-sample gain}
\label{sec:results-keep}

\begin{figure}[t]
    \centering
    \includegraphics[width=\linewidth]{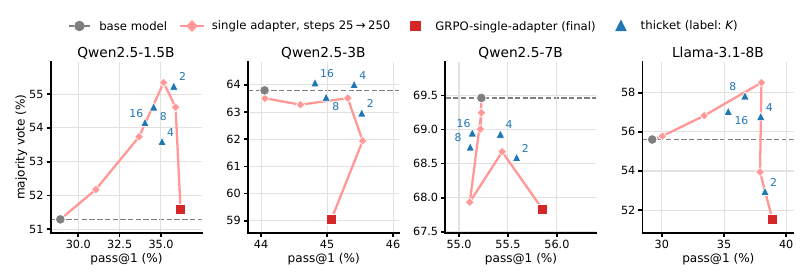}
    \caption{\textbf{Single-sample accuracy vs.\ majority vote} on the math suite (pass@1 = mean voter accuracy; majority vote over the fixed $160$-completion pool; macro means, mean over three seeds). The pink path follows the single adapter from the untrained base (gray circle) through steps $25/50/125/250$ to its final checkpoint, \single{} (red square). These checkpoints are the step-matched pools for $K = 16/8/4/2$. Blue triangles are thickets, labeled by $K$. Dashed line: the base model's vote. Full training moves right but, on three of four models, down below the base. Thickets with $K{\geq}4$ keep most of the rightward move where there is one, while staying within $0.8$ p.p.\ of the base's vote or above it. At $K \leq 4$ the matched checkpoints sit at about the thickets' pass@1, and the vote goes either way. At $K \geq 8$ the thickets lie to the right of and above the matched checkpoints on three of four models.}
    \label{fig:tradeoff}
\end{figure}

Avoiding the damage is easy if we give up training: the untrained base does it, and so, at the right step, does stopping the single run early. The fair question is whether a thicket keeps more of RLVR's single-sample gain than a single adapter stopped at a member's step count. The pink path in Figure~\ref{fig:tradeoff} is that control.

At small $K$, early stopping is as good as splitting. On Qwen2.5-1.5B and Llama, where full training lifts pass@1 by $+7.3$ and $+9.7$ p.p.\ (only $+1.0$ and $+0.7$ on Qwen2.5-3B and 7B, too little to divide), thicket members keep $94$--$95\%$ of that gain at $K{=}2$ and $85$--$91\%$ at $K{=}4$. The step-matched adapter keeps about as much ($90$--$96\%$ and $86$--$91\%$). At $K{=}4$ it also votes higher, $+4.0$ and $+2.9$ p.p.\ above the base against $+2.3$ and $+1.2$ for the thicket. Across all four models at $K \leq 4$, the two win four of eight vote comparisons each.

At large $K$, splitting wins clearly. The step-matched run keeps only $43$--$66\%$ of the pass@1 gain at $K{=}8$ and $8$--$30\%$ at $K{=}16$ on the same two models. Thicket members, each trained on $\nicefrac{1}{8}$ or $\nicefrac{1}{16}$ of the budget and its own shard, keep $77$--$78\%$ and $63$--$70\%$. Their voters are $0.9$ to $5.3$ p.p.\ more accurate (on three of four models overall), and the thicket out-votes the step-matched pool by $+0.9$ to $+2.0$ p.p. Across all four models at $K \geq 8$, the thicket wins six of eight comparisons, ties one, and loses one (Qwen2.5-7B at $K{=}8$, by $0.5$).

Splitting is also the more robust choice. The thicket's vote varies by at most $1.0$ p.p.\ across $K \in \{4, 8, 16\}$ on every model, whereas the step-matched vote varies by more than $1.0$ p.p.\ on three of four models (up to $3.1$). The single run's best step also differs across models (Section~\ref{sec:results-dyn}).

Why does the thicket keep the accuracy but not the damage? Its members are competent without being redundant. At $K{=}2$ they are on average as accurate as the fully trained adapter (within $0.6$ p.p.\ on every model's math suite), yet thicket voters are less error-correlated than \single{} voters in all $32$ (model, suite, $K$) settings, and at $K{=}2$ the thicket out-votes \single{} by $+0.8$ to $+4.0$ p.p. At $K{=}16$ they add at most $+0.07$ correlation over the base, against $+0.05$ to $+0.20$ for \single{}. Appendix~\ref{app:corr} splits the thicket's margin over \single{} into the RLVR gain it keeps and the vote damage it avoids (Table~\ref{tab:decomp}).

\subsection{The cost of concentration grows with the vote budget}
\label{sec:budget}

\begin{figure}[t]
    \centering
    \includegraphics[width=0.66\linewidth]{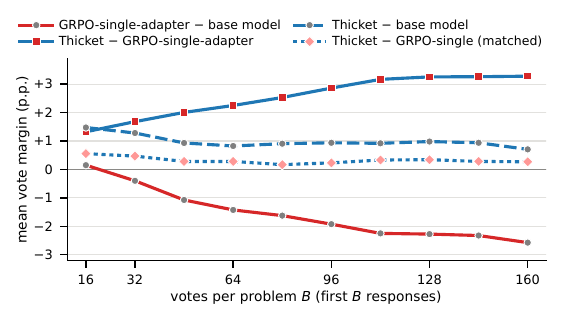}
    \caption{\textbf{The cost of concentration grows with the vote budget.} Mean majority-vote margin on the math suite (p.p.) over the $16$ (model, $K$) settings, as every method votes over its first $B$ completions in generation order. Each setting's margin is the difference of displayed three-seed means. Line colour marks the method in front, and markers mark the method subtracted. Red line with gray circles: \single{} $-$ base. Blue line with red squares: thicket $-$ \single{}. Dashed blue with gray circles: thicket $-$ base. Dotted blue with pink diamonds: thicket $-$ \singlem{}. $B{=}160$ is Table~\ref{tab:mv}. Table~\ref{tab:budget} in Appendix~\ref{app:budget} gives the values.}
    \label{fig:budget}
\end{figure}

We now ask how the comparison changes with the number of votes. For $B = 16, 32, \dots, 160$, every method votes over its first $B$ completions in generation order (Appendix~\ref{app:eval}), so smaller budgets are prefixes of larger ones. A thicket member contributes its first $\nicefrac{B}{K}$ completions, one each at $B{=}16$ and $K{=}16$. A single policy contributes the completions drawn with its first $B$ sampling seeds, so the base and \single{} votes are again identical at every $K$. Figure~\ref{fig:budget} reports mean margins over the $16$ (model, $K$) settings.

The fully trained adapter's deficit grows with the budget. At $16$ votes it is level with the base on average ($+0.15$ p.p.). It votes above the base on Qwen2.5-1.5B and Llama ($+1.4$ and $+1.8$ p.p.) and below it on Qwen2.5-3B and 7B ($-1.5$ and $-1.1$). From $32$ votes on it trails the base on average, and the gap widens to $2.58$ p.p.\ at $160$ votes. On Llama it leads the base by $1.8$ p.p.\ at $16$ votes and trails it by $4.1$ at $160$. Extra samples help its vote much less. From $16$ to $160$ votes its majority vote rises by $3.78$ p.p.\ on average, against $6.50$ for the base, $6.02$ for \singlem{}, and $5.73$ for the thicket (means of the per-setting gains). Its voters have the most correlated errors of the four pools in all $32$ (model, suite, $K$) settings (Appendix~\ref{app:corr}). This is consistent with extra samples repeating answers the pool has already given.

The thicket's mean lead over \single{} never shrinks as the budget grows. It rises from $+1.33$ p.p.\ at $16$ votes (ahead in $15$ of $16$ settings) to $+3.28$ at $160$ (ahead in all $16$). The growth comes from three models. On Qwen2.5-7B, where RLVR adds least, the lead stays between $+0.78$ and $+1.18$ p.p. The thicket's leads over the other two controls shrink. Over the base, the mean lead falls from $+1.48$ p.p.\ at $16$ votes (ahead in $13$ settings) to between $+0.83$ and $+0.98$ from $48$ to $144$ votes, and to $+0.71$ at $160$ (ahead in $9$). Over \singlem{}, it falls from $+0.56$ (ahead in $11$) to $+0.27$ (ahead in $10$). Averaged over $K$, each trained pool gains less from extra votes than the base, so each loses ground to it as the budget grows.

Within the thicket, the loss again follows concentration. A larger $K$ spreads the budget more thinly, and the thicket's vote gains more from extra samples: $+4.68$ p.p.\ from $16$ to $160$ votes at $K{=}2$, $+5.43$ at $K{=}4$, $+6.35$ at $K{=}8$, and $+6.48$ at $K{=}16$, level with the base's $+6.50$. From $16$ to $160$ votes the thicket's mean lead over the base falls from $+1.73$ to $-0.10$ p.p.\ at $K{=}2$ and from $+1.85$ to $+0.78$ at $K{=}4$, but only from $+1.28$ to $+1.13$ at $K{=}8$ and from $+1.05$ to $+1.03$ at $K{=}16$. Appendix~\ref{app:budget} plots every curve and repeats Tables~\ref{tab:mv} and~\ref{tab:cov} at $16$ votes.

\subsection{Confidence-filtered voting}
\label{sec:results-lik}

A pool that has lost its dissent should also gain little from weighing its votes. We test this with \emph{confidence-filtered voting}. Each voter keeps only its top-$q$ fraction of completions by mean log-likelihood $\overline{\ell}$, and plain majority voting runs over the survivors. We select $q \in \{1, \nicefrac{1}{2}, \nicefrac{1}{4}, \text{top-}1\}$ by per-benchmark five-fold cross-validation, so every problem is scored under a $q$ chosen without seeing it, and $q{=}1$ (plain voting) is always available (Appendix~\ref{app:eval}). We score all four methods at $160$ votes, each against its own plain majority vote. Correct completions have the higher median $\overline{\ell}$ in every method--model pool at $K{=}16$ (Figure~\ref{fig:loglik_sep} in Appendix~\ref{app:mv}), so the filter has a signal to use.

Filtering helps three of the four pools, and it gives nothing to the pool whose errors are most correlated. Its mean gain over plain voting is $+0.53 \pm 0.10$ p.p.\ for the thicket, $+0.51 \pm 0.06$ for the untrained base, and $+0.43 \pm 0.08$ for \singlem{} (means over the $16$ settings of differences of displayed three-seed means; $\pm$ std over seeds). The thicket gains in $15$ of $16$ settings (by up to $1.1$ p.p.), and the base and \singlem{} in all $16$. For the fully trained \single{} it is $+0.01 \pm 0.14$, that is, no gain. Filtering raises its vote in $6$ settings, leaves it unchanged in $4$, and lowers it in $6$. We read this as a sharpened policy leaving little dissent to filter. Filtering widens the thicket's lead over \single{} from $+3.28$ to $+3.80$ p.p.\ and leaves its lead over the base almost unchanged ($+0.71$ to $+0.73$).

\subsection{Random shards are enough}
\label{sec:subject}

\begin{table}[t]
  \caption{\textbf{Subject-specialist vs.\ random disjoint shards at $K{=}7$} ($184$ training problems per adapter in both arms; $7 \times 23 = 161$ completions per problem; single seed, seed $42$). Macro means (\%) over the five math benchmarks, except $\corr$. Bold: better arm per model and column (ties bold both; $\corr$ is not bolded).}
  \label{tab:subject}
  \begin{center}
  \footnotesize
  \setlength{\tabcolsep}{4pt}
  \begin{tabular}{llccccc}
\toprule
Model & Shards & Majority vote & Filtered vote & Coverage & Voter acc. & $\corr$ \\
\midrule
Qwen2.5-1.5B & Subject & 54.2 & 55.2 & \textbf{89.6} & \textbf{34.4} & 0.37 \\
 & Random & \textbf{54.8} & \textbf{56.0} & \textbf{89.6} & \textbf{34.4} & 0.37 \\
\addlinespace[2pt]
Qwen2.5-3B & Subject & 62.9 & \textbf{63.4} & \textbf{91.5} & 44.8 & 0.44 \\
 & Random & \textbf{63.1} & 62.7 & 91.0 & \textbf{44.9} & 0.44 \\
\addlinespace[2pt]
Qwen2.5-7B & Subject & 69.1 & 69.3 & \textbf{92.7} & \textbf{55.2} & 0.51 \\
 & Random & \textbf{69.3} & \textbf{69.5} & 92.1 & 55.0 & 0.51 \\
\addlinespace[2pt]
Llama-3.1-8B & Subject & 57.1 & 57.9 & 89.1 & 35.9 & 0.34 \\
 & Random & \textbf{57.9} & \textbf{59.0} & \textbf{89.6} & \textbf{36.1} & 0.35 \\
\bottomrule
\end{tabular}

  \end{center}
\end{table}

Random disjoint sharding is only the simplest way to split a budget. A natural question is whether \emph{engineered} shards do better. We partition the MATH training pool by its subject metadata into seven specialist shards (Algebra, Counting \& Probability, Geometry, Intermediate Algebra, Number Theory, Prealgebra, Precalculus), equalized at $184$ examples per adapter (the size of the smallest subject). The control is a thicket of seven \emph{random} disjoint shards of the same size. Both arms use the same GRPO recipe and draw $23$ samples per member, for $161$ completions per problem. We run this ablation for one seed (Table~\ref{tab:subject}; Appendix~\ref{app:subject}).

Engineering the shards buys nothing. Random shards vote at least as well as subject shards on all four models ($+0.2$ to $+0.8$ p.p.), and win under confidence filtering on three of four. Coverage, voter accuracy, and error correlation are essentially identical across the two arms (within $0.6$ p.p., $0.2$ p.p., and $0.01$). Which disjoint split a thicket uses matters little, and random shards are enough.

\section{Discussion, limitations, and conclusion}
\label{sec:discussion}

\textbf{Practical picture.} The standard pipeline trains one policy with RLVR and then samples it many times. Our results say this composition is lossy. Training raises pass@1, but it spends the diversity the vote needs, and at a fixed inference budget the vote can end below where it started. The more votes a user draws, the larger the loss. Splitting the budget is a drop-in fix that changes how training is spent, not how inference works. At deployment a thicket is $K$ LoRA adapters over one frozen base model, and modern multi-adapter servers batch concurrent LoRAs over shared base weights at near-single-model cost~\citep{sheng2023slora}. Training is embarrassingly parallel: on $K$ workers a thicket finishes in $\nicefrac{1}{K}$ of the single adapter's wall-clock time. The recipe adds no hyperparameters beyond $K$, and for $K \geq 4$ the choice of $K$ is forgiving (Section~\ref{sec:results-keep}). The two-member thicket is the least reliable configuration: it is the weakest thicket on three of four models and has the largest seed spread of any thicket entry in Table~\ref{tab:mv}.

\textbf{Limitations.} Our study covers one RL algorithm (GRPO), voter counts $K \leq 16$, models up to $8$B, and budgets of $16$ to $160$ votes. Thickets protect the vote rather than guarantee a gain over the base. Their lead over the base and over \singlem{} shrinks as the budget grows, and at $160$ votes they beat the base in $9$ of $16$ settings. On Llama the base's vote is still rising at $160$ votes, so the lead may shrink further at larger budgets. At a single sample the order reverses: \single{} has the highest pass@1 of the four pools on Qwen2.5-1.5B, Qwen2.5-7B, and Llama at every $K$. When RLVR adds little pass@1, as on Qwen2.5-7B, a thicket roughly matches the untrained model. All arms train without a KL penalty (Table~\ref{tab:hyper}), and we do not test whether KL regularization changes the single adapter's vote. Majority voting needs comparable answers, so our voting results are math-only, and code enters only through coverage, pass@1, and correlation. The subject-shard ablation uses a single seed. Finally, we test only random and subject-based sharding.

\textbf{Conclusion.} Majority voting and RLVR are each hard to beat, but composing them naively is costly. Concentrating an RLVR budget in one policy spends the diversity the vote needs: on three of four models its vote ends below the untrained model's, by more as votes grow. Splitting the same budget across disjoint shards keeps most of the accuracy and avoids the damage. When the plan is to sample and vote, train a thicket, not a champion.\label{endofmaintext}

\clearpage
\appendix
% appendix float placement: let small floats share pages instead of one float per page
\renewcommand{\topfraction}{0.9}\renewcommand{\bottomfraction}{0.8}\renewcommand{\textfraction}{0.07}
\renewcommand{\floatpagefraction}{0.5}\setcounter{topnumber}{3}\setcounter{bottomnumber}{2}\setcounter{totalnumber}{4}

\section{Training details}
\label{app:hyperparams}

\textbf{Models.} Qwen2.5-1.5B-Instruct, Qwen2.5-3B-Instruct, Qwen2.5-7B-Instruct~\citep{yang2024qwen25}, and Llama-3.1-8B-Instruct~\citep{grattafiori2024llama3}, all frozen; only LoRA parameters train.

\textbf{Data and partitioning.} All math experiments use a fixed pool of $D=2{,}000$ problems sampled from the MATH training split; all code experiments use a fixed pool of $D=2{,}000$ problems from the LeetCodeDataset training split. For a thicket of size $K$, the pool is partitioned uniformly at random (partition seed $\in \{42, 8, 23\}$) into $K$ disjoint shards of exactly $\nicefrac{2000}{K}$ problems. Shards are sampled without replacement, so the union of the $K$ shards is exactly the single adapter's training set.

\textbf{Seeds.} Each seed fixes the thicket's data partition, the data ordering (shuffle) for the single-adapter runs, and the sampling seeds of every evaluation pool; adapter initialization seeds are held fixed, so reported stds isolate data-allocation and sampling randomness.

\textbf{GRPO training and checkpoints.} All runs use the same verl-based GRPO implementation~\citep{sheng2025hybridflow,shao2024deepseekmath} with group-normalized advantages, a binary verified reward, and identical hyperparameters (Table~\ref{tab:hyper}); the \emph{only} difference between arms is the training file. The single-adapter runs train for $\approx 500$ optimizer steps and checkpoint every $25$; the \singlem{} control evaluates the last saved checkpoint at or before a thicket member's step count (steps $250/125/50/25$ for $K{=}2/4/8/16$, where members train $248$--$250$, $124$--$125$, $62$--$63$, and $31$--$32$ steps), and Figure~\ref{fig:dynamics} evaluates steps $\{25, 50, 125, 250, \text{final}\}$. Thicket member $i$ trains for one epoch on its shard ($\approx\nicefrac{500}{K}$ steps).

\begin{table}[!htbp]
  \caption{Shared GRPO/LoRA hyperparameters. The two training arms differ only in the training file.}
  \label{tab:hyper}
  \begin{center}
  \footnotesize
  \setlength{\tabcolsep}{4pt}
  \begin{tabular}{llll}
    \toprule
    LoRA rank / $\alpha$ / dropout & 64 / 128 / 0.0 & prompts per step & 4 \\
    LoRA targets & q,k,v,o,gate,up,down proj. & rollouts per prompt & 8 \\
    learning rate & $2 \times 10^{-5}$ (constant, 5\% warmup) & rollout temperature & 1.0 \\
    weight decay / grad.\ clip & 0.01 / 0.1 & max prompt / completion & 1{,}024 / 2{,}048 tok. \\
    KL coefficient & 0.0 & reward & binary verified \\
    advantage & group-normalized (GRPO) & epochs & 1 \\
    \bottomrule
  \end{tabular}
  \end{center}
\end{table}

\section{Evaluation protocol}
\label{app:eval}

Zero-shot, temperature $1.0$, top-$p$ $0.95$, max $2{,}048$ new tokens. \emph{Math:} MATH500 ($500$ problems), GSM8K ($1{,}319$), OlympiadBench ($674$), Countdown ($150$), Knights~\&~Knaves ($300$). \emph{Code:} MBPP ($427$), HumanEval ($164$), LeetCodeDataset ($228$). Suite-level numbers are macro means of per-benchmark values, computed per seed and then averaged over the three seeds (except the $K{=}7$ ablation of Appendix~\ref{app:subject}, seed $42$).

\textbf{The fixed $160$-vote budget.} Every (method, $K$) pool contains exactly $160$ completions per problem, so all comparisons -- across methods \emph{and} across $K$ -- hold the inference budget fixed. A thicket fields $K$ members, and each member contributes $\nicefrac{160}{K}$ completions per problem ($80/40/20/10$ for $K{=}2/4/8/16$). A single policy -- the base model, the final \single{} checkpoint, or the \singlem{} checkpoint (step $250/125/50/25$ for $K{=}2/4/8/16$) -- is sampled once as a pool of $160$ completions per problem, drawn as $10$ requests of $16$ parallel samples, and the $K$-voter view assigns these i.i.d.\ completions to $K$ nominal voters of $\nicefrac{160}{K}$ each. The majority vote and coverage of the base model and \single{} are therefore identical at every $K$ (we verified this to machine precision), and only the pairwise error-correlation estimate depends, negligibly, on the grouping. Every completion in every pool is drawn with its own sampling seed.

\textbf{The first-$B$ convention.} Section~\ref{sec:budget} and Appendix~\ref{app:budget} vote over a prefix of each pool in generation order. At a budget of $B \in \{16, 32, \dots, 160\}$ votes, each thicket member contributes its first $\nicefrac{B}{K}$ completions ($8/4/2/1$ at $B{=}16$ for $K{=}2/4/8/16$). A single policy contributes the completions drawn with its first $B$ sampling seeds, and the $K$-voter view assigns them to $K$ nominal voters. The majority vote and coverage of the base model and \single{} are therefore identical across $K$ at every $B$ (we verified this to machine precision). Every $B$ is a multiple of $16$, so $\nicefrac{B}{K}$ is a whole number at every $K$, and $B{=}160$ is the full pool of the main tables. Each budget is one real deployment of that many samples, and smaller budgets are prefixes of larger ones.

\textbf{Voting details.} Completions from which no candidate answer can be parsed (\emph{nulls}) are dropped before voting. Ties among plurality answers are scored exactly as the expected accuracy under uniform tie-breaking, removing tie-break randomness. Coverage counts a problem as covered if \emph{any} pooled completion is graded correct.

\textbf{Answer canonicalization.} Votes are cast over canonical equivalence classes of answers rather than exact strings. On MATH500, GSM8K, and OlympiadBench we parse each extracted answer with \texttt{math\_verify}: numeric answers are keyed by value (to $12$ significant digits, so $0.5 \equiv \nicefrac{1}{2} \equiv$ comma variants), and non-numeric answers (tuples, intervals, symbolic expressions) by their parse tree; unparseable strings fall back to a whitespace/LaTeX-normalized key. On Countdown we normalize notation only ($\times \equiv *$, $\div \equiv /$, spacing) -- structurally different expressions are different solutions -- and on Knights~\&~Knaves the assignment is canonicalized order-insensitively. A canonical candidate is graded correct if any member completion was graded correct by the verifier.

\textbf{Confidence-filtered voting details.} $\overline{\ell}(r)$ is the mean per-token log-likelihood of response $r$ under the model or adapter that generated it (available for math; code pools are not likelihood-scored). All four arms are scored on all three seeds at $160$ votes, each against its own plain majority vote over the same completions. Within each problem, each voter ranks its own $\nicefrac{160}{K}$ completions by $\overline{\ell}$ and keeps the top $q$ fraction (at least one), with $q \in \{1, \nicefrac{1}{2}, \nicefrac{1}{4}, \text{top-}1\}$. Plain majority voting then runs over the surviving completions under the usual canonical-class, null-dropped, tie-marginalized protocol. $q$ is selected per benchmark on four folds and applied to the held-out fold (fold $=$ hash of the problem key). Exact training-fold ties are resolved toward the weaker filter, so $q{=}1$ (no filtering) is always available as a fallback. Because each nominal voter filters its own completions, the filtered vote of the base model and \single{}, unlike their plain vote, depends on $K$. Across the $1{,}200$ selections per arm ($3$ seeds $\times$ $16$ settings $\times$ $5$ benchmarks $\times$ $5$ folds), the CV picks a real filter ($q \neq 1$) in $67\%$ of cases for the base model, $64\%$ for the thicket, $59\%$ for \singlem{}, and $53\%$ for \single{}. It picks the aggressive top-$1$-per-voter setting in at most $2.5\%$ of any arm's selections ($0.8\%$ for the thicket). Confidence is informative, but the vote still wants breadth. Figure~\ref{fig:loglik_sep} visualizes the correct-vs-incorrect likelihood separation the filter exploits.

\section{Per-benchmark majority vote and confidence-filtered voting}
\label{app:mv}

\begin{figure}[!htbp]
    \centering
    \includegraphics[width=0.95\linewidth]{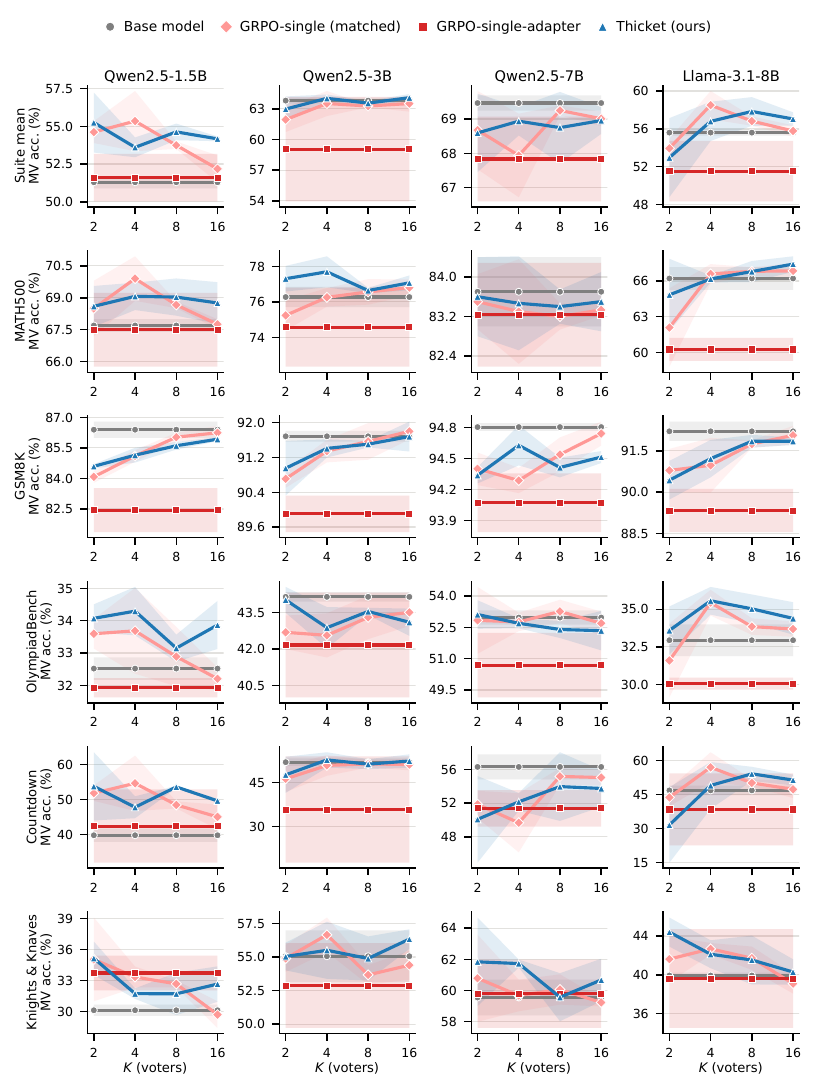}
    \caption{Majority-vote accuracy vs.\ voter count $K$ (fixed $160$-vote budget at every $K$): suite macro (top row) and each benchmark, all models (columns). Lines: seed means; bands: $\pm 1$ std. Base and \single{} are flat by construction (one $160$-completion pool regrouped at every $K$).}
    \label{fig:scaling_mv_math}
\end{figure}

\begin{figure}[!htbp]
    \centering
    \includegraphics[width=\linewidth]{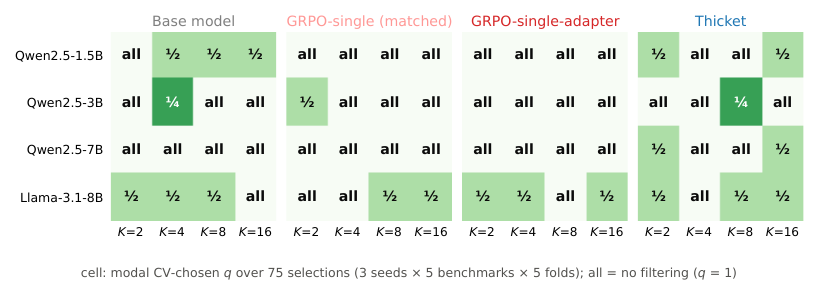}
    \caption{Most frequently selected CV filter strength per (model, $K$) setting and method (mode over $75$ selections: three seeds $\times$ five math benchmarks $\times$ five folds; ties toward the weaker filter, consistent with the selection rule; darker $=$ more aggressive filtering). The most frequent choice is no filtering in $13$ of $16$ settings for \single{} and for \singlem{}, in $9$ for the base model, and in $8$ for the thicket. Because real-filter selections split between $\nicefrac{1}{2}$ and $\nicefrac{1}{4}$, the mode understates how often each pool filters. Across all $1{,}200$ selections per method, the CV picks a real filter in $67\%$ (base), $64\%$ (thicket), $59\%$ (\singlem{}), and $53\%$ (\single{}) of cases.}
    \label{fig:topm_modes}
\end{figure}

\begin{figure}[!htbp]
    \centering
    \includegraphics[width=0.95\linewidth]{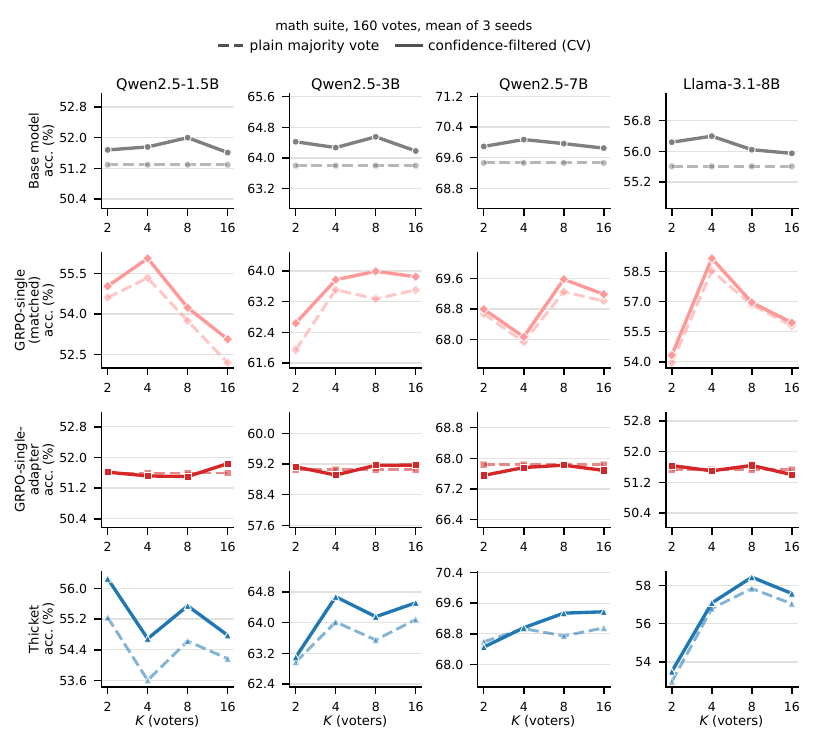}
    \caption{Plain majority vote (dashed) vs.\ confidence-filtered voting (solid) on the math-suite macro at $160$ votes, per method (rows) and model (columns); mean of three seeds. By displayed means, the solid curve sits above the dashed one in all $16$ settings for the base model and \singlem{} and in $15$ of $16$ for the thicket. For \single{} it is above in $6$ settings, level in $4$, and below in $6$. No bands are drawn, because bands of the two curves would hide a gain of about half a point. Tables~\ref{tab:lik}--\ref{tab:lik_singlem} give the stds.}
    \label{fig:lik_vs_mv}
\end{figure}

\begin{figure}[!htbp]
    \centering
    \includegraphics[width=0.9\linewidth]{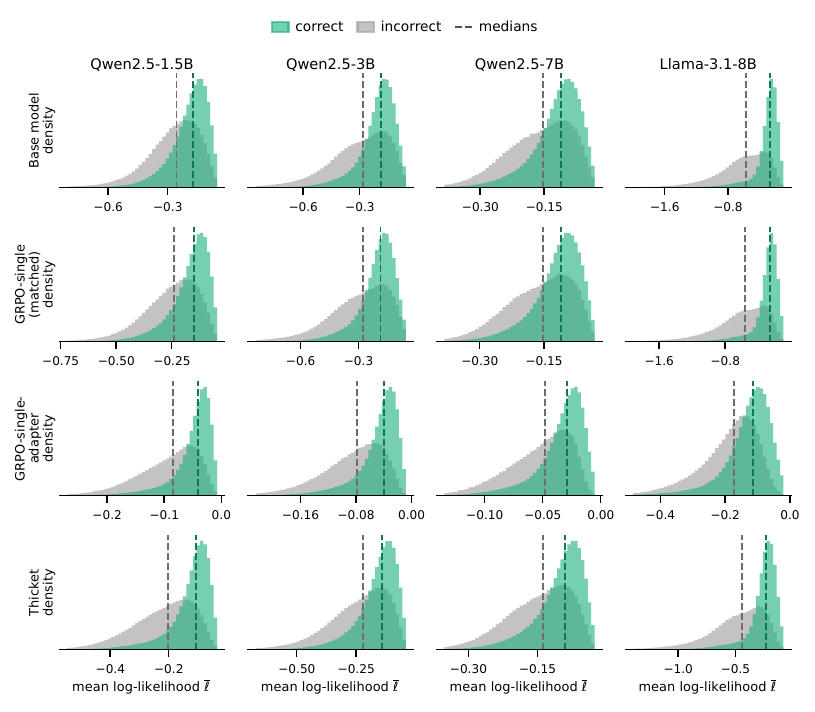}
    \caption{\textbf{Why confidence is worth acting on.} Distributions of the per-response mean log-likelihood $\overline{\ell}$ for correct (green) vs.\ incorrect (gray) completions at $K{=}16$, over all $160$ completions per problem, pooled over the three seeds and the five math benchmarks (non-null responses; histogram ranges clipped to the $0.5$--$99.5$th percentiles; dashed lines: medians, dark green for correct and gray for incorrect). x-axes differ across panels. Correct completions have the higher median in every method--model pool -- the signal confidence-filtered voting exploits. The fully trained \single{} (third row) concentrates its completions much closer to $\overline{\ell}{=}0$ than the other pools.}
    \label{fig:loglik_sep}
\end{figure}

Tables~\ref{tab:mv_perbench} and~\ref{tab:mv_perbench_b} break the majority vote of Table~\ref{tab:mv} down by benchmark, and Tables~\ref{tab:lik_perbench} and~\ref{tab:lik_perbench_b} do the same for confidence-filtered voting. Figure~\ref{fig:scaling_mv_math} plots the majority-vote comparison against voter count $K$ (at a fixed $160$ votes) with $\pm 1$ std bands. Table~\ref{tab:lik} gives the thicket's confidence-filtered grid, and Tables~\ref{tab:lik_base}--\ref{tab:lik_singlem} replicate its format for the base model, \single{}, and \singlem{}. Figure~\ref{fig:lik_vs_mv} compares plain and confidence-filtered voting for all four methods, Figure~\ref{fig:topm_modes} shows the most frequent CV-selected filter strength per setting, and Figure~\ref{fig:loglik_sep} shows the correct-vs-incorrect likelihood separation the filter exploits. All likelihood results are three-seed means, with the sample standard deviation after $\pm$.

\begin{figure}[!htbp]
    \centering
    \includegraphics[width=\linewidth]{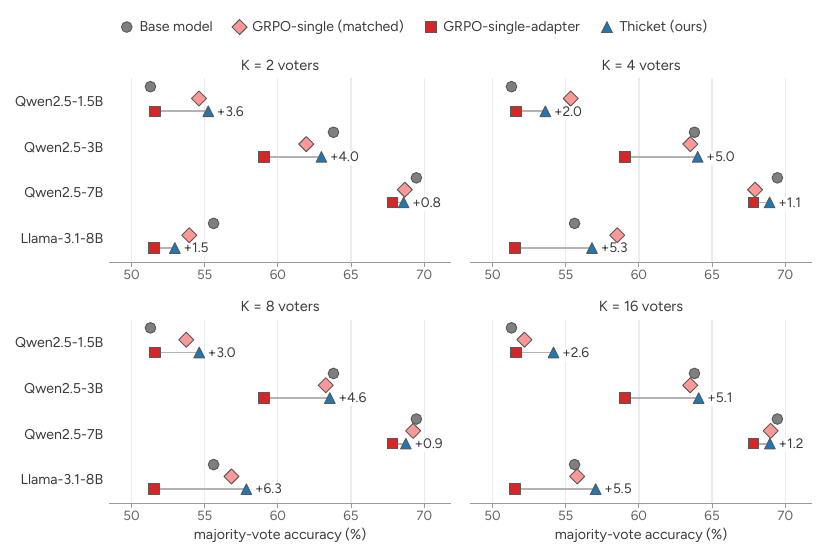}
    \caption{\textbf{The headline comparison at every voter count.} Per-model majority-vote accuracy on the math suite (macro mean over five benchmarks, mean over three seeds) for all four methods, at $K \in \{2,4,8,16\}$. The $K{=}16$ panel reproduces the right panel of Figure~\ref{fig:schematic}. Number at the end of each red--blue connector: thicket $-$ \single{} accuracy.}
    \label{fig:headline_grid}
\end{figure}

Figure~\ref{fig:headline_grid} reproduces the headline comparison of Figure~\ref{fig:schematic} (right) --- per-model majority-vote accuracy for all four methods on the math suite --- at every voter count $K$, not only $K{=}16$.

\begin{table}[!htbp]
  \caption{Per-benchmark majority-vote accuracy (\%) at $K \in \{2,4\}$, fixed $160$-vote pools; mean\,$\pm$\,std over three seeds. Mean rows are macro means. Bold: best method per row and $K$; underlined: second best.}
  \label{tab:mv_perbench}
  \begin{center}
  \scriptsize
  \setlength{\tabcolsep}{2.5pt}
  \begin{tabular}{llcccccccc}
\toprule
 & & \multicolumn{2}{c}{Base} & \multicolumn{2}{c}{\singlem{}} & \multicolumn{2}{c}{\single{}} & \multicolumn{2}{c}{Thicket} \\
\cmidrule(lr){3-4} \cmidrule(lr){5-6} \cmidrule(lr){7-8} \cmidrule(lr){9-10}
Model & Benchmark ($K{=}$) & 2 & 4 & 2 & 4 & 2 & 4 & 2 & 4 \\
\midrule
Qwen2.5-1.5B & MATH500 & 67.7\eb{0.3} & 67.7\eb{0.3} & \underline{68.5\eb{1.3}} & \textbf{69.9\eb{1.0}} & 67.5\eb{1.7} & 67.5\eb{1.7} & \textbf{68.6\eb{1.0}} & \underline{69.1\eb{0.6}} \\
 & GSM8K & \textbf{86.4\eb{0.4}} & \textbf{86.4\eb{0.4}} & 84.1\eb{0.3} & \underline{85.1\eb{0.3}} & 82.4\eb{1.1} & 82.4\eb{1.1} & \underline{84.6\eb{0.1}} & \underline{85.1\eb{0.4}} \\
 & OlympiadBench & 32.5\eb{0.4} & 32.5\eb{0.4} & \underline{33.6\eb{0.5}} & \underline{33.7\eb{1.3}} & 31.9\eb{0.3} & 31.9\eb{0.3} & \textbf{34.1\eb{0.4}} & \textbf{34.3\eb{0.8}} \\
 & Countdown & 39.8\eb{1.9} & 39.8\eb{1.9} & \underline{51.8\eb{2.2}} & \textbf{54.7\eb{8.0}} & 42.4\eb{10.5} & 42.4\eb{10.5} & \textbf{53.8\eb{9.8}} & \underline{47.8\eb{3.2}} \\
 & Knights \& Knaves & 30.1\eb{0.5} & 30.1\eb{0.5} & \underline{35.0\eb{4.1}} & \underline{33.3\eb{1.0}} & 33.7\eb{1.7} & \textbf{33.7\eb{1.7}} & \textbf{35.1\eb{1.6}} & 31.7\eb{0.5} \\
 & \emph{Mean} & 51.3\eb{0.4} & 51.3\eb{0.4} & \underline{54.6\eb{0.7}} & \textbf{55.3\eb{2.0}} & 51.6\eb{1.6} & 51.6\eb{1.6} & \textbf{55.2\eb{2.0}} & \underline{53.6\eb{0.7}} \\
\addlinespace[2pt]
Qwen2.5-3B & MATH500 & \underline{76.3\eb{0.6}} & \underline{76.3\eb{0.6}} & 75.2\eb{0.9} & \underline{76.3\eb{0.5}} & 74.6\eb{2.2} & 74.6\eb{2.2} & \textbf{77.3\eb{0.7}} & \textbf{77.7\eb{0.9}} \\
 & GSM8K & \textbf{91.7\eb{0.1}} & \textbf{91.7\eb{0.1}} & 90.7\eb{0.2} & 91.3\eb{0.3} & 89.9\eb{0.4} & 89.9\eb{0.4} & \underline{91.0\eb{0.6}} & \underline{91.4\eb{0.2}} \\
 & OlympiadBench & \textbf{44.1\eb{0.2}} & \textbf{44.1\eb{0.2}} & 42.7\eb{0.7} & 42.6\eb{0.9} & 42.2\eb{2.1} & 42.2\eb{2.1} & \underline{44.0\eb{0.5}} & \underline{42.9\eb{0.8}} \\
 & Countdown & \textbf{51.9\eb{1.3}} & \underline{51.9\eb{1.3}} & 46.2\eb{4.5} & 50.8\eb{3.5} & 35.7\eb{18.1} & 35.7\eb{18.1} & \underline{47.6\eb{6.1}} & \textbf{52.6\eb{2.6}} \\
 & Knights \& Knaves & \textbf{55.1\eb{1.9}} & 55.1\eb{1.9} & 54.9\eb{1.0} & \textbf{56.6\eb{1.3}} & 52.9\eb{3.2} & 52.9\eb{3.2} & \underline{55.0\eb{1.0}} & \underline{55.5\eb{2.1}} \\
 & \emph{Mean} & \textbf{63.8\eb{0.3}} & \underline{63.8\eb{0.3}} & 61.9\eb{1.2} & 63.5\eb{1.2} & 59.0\eb{5.1} & 59.0\eb{5.1} & \underline{63.0\eb{1.2}} & \textbf{64.0\eb{0.4}} \\
\addlinespace[2pt]
Qwen2.5-7B & MATH500 & \textbf{83.7\eb{0.7}} & \textbf{83.7\eb{0.7}} & 83.5\eb{0.6} & 83.3\eb{1.1} & 83.2\eb{1.1} & 83.2\eb{1.1} & \underline{83.6\eb{0.8}} & \underline{83.5\eb{1.0}} \\
 & GSM8K & \textbf{94.8\eb{0.0}} & \textbf{94.8\eb{0.0}} & \underline{94.4\eb{0.2}} & 94.3\eb{0.1} & 94.1\eb{0.3} & 94.1\eb{0.3} & 94.3\eb{0.1} & \underline{94.6\eb{0.2}} \\
 & OlympiadBench & \underline{53.0\eb{0.3}} & \textbf{53.0\eb{0.3}} & 52.8\eb{1.6} & \underline{52.8\eb{0.4}} & 50.7\eb{1.5} & 50.7\eb{1.5} & \textbf{53.1\eb{0.7}} & 52.7\eb{0.2} \\
 & Countdown & \textbf{56.3\eb{1.5}} & \textbf{56.3\eb{1.5}} & \underline{51.9\eb{1.6}} & 49.7\eb{3.5} & 51.4\eb{2.2} & 51.4\eb{2.2} & 50.1\eb{5.1} & \underline{52.2\eb{1.0}} \\
 & Knights \& Knaves & 59.5\eb{1.1} & 59.5\eb{1.1} & \underline{60.8\eb{2.8}} & 59.7\eb{1.0} & 59.8\eb{2.2} & \underline{59.8\eb{2.2}} & \textbf{61.8\eb{2.8}} & \textbf{61.7\eb{0.3}} \\
 & \emph{Mean} & \textbf{69.5\eb{0.2}} & \textbf{69.5\eb{0.2}} & \underline{68.7\eb{1.1}} & 67.9\eb{1.2} & 67.8\eb{1.2} & 67.8\eb{1.2} & 68.6\eb{1.1} & \underline{68.9\eb{0.4}} \\
\addlinespace[2pt]
Llama-3.1-8B & MATH500 & \textbf{66.2\eb{1.0}} & \underline{66.2\eb{1.0}} & 62.1\eb{3.3} & \textbf{66.6\eb{0.8}} & 60.3\eb{1.0} & 60.3\eb{1.0} & \underline{64.8\eb{3.0}} & \underline{66.2\eb{0.3}} \\
 & GSM8K & \textbf{92.2\eb{0.4}} & \textbf{92.2\eb{0.4}} & \underline{90.8\eb{0.8}} & 91.0\eb{1.0} & 89.3\eb{0.8} & 89.3\eb{0.8} & 90.4\eb{0.7} & \underline{91.2\eb{0.7}} \\
 & OlympiadBench & \underline{32.9\eb{1.1}} & 32.9\eb{1.1} & 31.6\eb{2.5} & \underline{35.4\eb{0.9}} & 30.0\eb{0.4} & 30.0\eb{0.4} & \textbf{33.6\eb{1.6}} & \textbf{35.6\eb{0.9}} \\
 & Countdown & \textbf{46.8\eb{2.1}} & 46.8\eb{2.1} & \underline{43.7\eb{3.9}} & \textbf{56.9\eb{6.6}} & 38.4\eb{15.8} & 38.4\eb{15.8} & 31.6\eb{16.6} & \underline{48.9\eb{10.3}} \\
 & Knights \& Knaves & 39.9\eb{0.3} & 39.9\eb{0.3} & \underline{41.6\eb{2.4}} & \textbf{42.7\eb{0.6}} & 39.6\eb{5.1} & 39.6\eb{5.1} & \textbf{44.4\eb{1.5}} & \underline{42.1\eb{1.4}} \\
 & \emph{Mean} & \textbf{55.6\eb{0.3}} & 55.6\eb{0.3} & \underline{53.9\eb{1.0}} & \textbf{58.5\eb{1.5}} & 51.5\eb{3.2} & 51.5\eb{3.2} & 53.0\eb{4.2} & \underline{56.8\eb{2.1}} \\
\bottomrule
\end{tabular}

  \end{center}
\end{table}

\begin{table}[!htbp]
  \caption{Per-benchmark majority-vote accuracy (\%) at $K \in \{8,16\}$, fixed $160$-vote pools (format of Table~\ref{tab:mv_perbench}).}
  \label{tab:mv_perbench_b}
  \begin{center}
  \scriptsize
  \setlength{\tabcolsep}{2.5pt}
  \begin{tabular}{llcccccccc}
\toprule
 & & \multicolumn{2}{c}{Base} & \multicolumn{2}{c}{\singlem{}} & \multicolumn{2}{c}{\single{}} & \multicolumn{2}{c}{Thicket} \\
\cmidrule(lr){3-4} \cmidrule(lr){5-6} \cmidrule(lr){7-8} \cmidrule(lr){9-10}
Model & Benchmark ($K{=}$) & 8 & 16 & 8 & 16 & 8 & 16 & 8 & 16 \\
\midrule
Qwen2.5-1.5B & MATH500 & 67.7\eb{0.3} & 67.7\eb{0.3} & \underline{68.7\eb{0.2}} & \underline{67.8\eb{0.2}} & 67.5\eb{1.7} & 67.5\eb{1.7} & \textbf{69.0\eb{0.9}} & \textbf{68.8\eb{1.0}} \\
 & GSM8K & \textbf{86.4\eb{0.4}} & \textbf{86.4\eb{0.4}} & \underline{86.0\eb{0.2}} & \underline{86.2\eb{0.3}} & 82.4\eb{1.1} & 82.4\eb{1.1} & 85.6\eb{0.2} & 85.9\eb{0.1} \\
 & OlympiadBench & 32.5\eb{0.4} & \underline{32.5\eb{0.4}} & \underline{32.9\eb{0.9}} & 32.2\eb{0.3} & 31.9\eb{0.3} & 31.9\eb{0.3} & \textbf{33.2\eb{0.4}} & \textbf{33.9\eb{0.7}} \\
 & Countdown & 39.8\eb{1.9} & 39.8\eb{1.9} & \underline{48.5\eb{1.0}} & \underline{45.0\eb{3.5}} & 42.4\eb{10.5} & 42.4\eb{10.5} & \textbf{53.6\eb{0.3}} & \textbf{49.6\eb{0.6}} \\
 & Knights \& Knaves & 30.1\eb{0.5} & 30.1\eb{0.5} & \underline{32.7\eb{0.3}} & 29.7\eb{1.3} & \textbf{33.7\eb{1.7}} & \textbf{33.7\eb{1.7}} & 31.7\eb{2.1} & \underline{32.6\eb{1.7}} \\
 & \emph{Mean} & 51.3\eb{0.4} & 51.3\eb{0.4} & \underline{53.7\eb{0.2}} & \underline{52.2\eb{0.9}} & 51.6\eb{1.6} & 51.6\eb{1.6} & \textbf{54.6\eb{0.5}} & \textbf{54.2\eb{0.2}} \\
\addlinespace[2pt]
Qwen2.5-3B & MATH500 & \underline{76.3\eb{0.6}} & 76.3\eb{0.6} & \textbf{76.6\eb{0.7}} & \underline{76.8\eb{0.5}} & 74.6\eb{2.2} & 74.6\eb{2.2} & \textbf{76.6\eb{0.2}} & \textbf{77.1\eb{0.4}} \\
 & GSM8K & \textbf{91.7\eb{0.1}} & \underline{91.7\eb{0.1}} & \underline{91.6\eb{0.4}} & \textbf{91.8\eb{0.2}} & 89.9\eb{0.4} & 89.9\eb{0.4} & 91.5\eb{0.1} & \underline{91.7\eb{0.3}} \\
 & OlympiadBench & \textbf{44.1\eb{0.2}} & \textbf{44.1\eb{0.2}} & 43.3\eb{0.9} & \underline{43.5\eb{0.6}} & 42.2\eb{2.1} & 42.2\eb{2.1} & \underline{43.5\eb{0.1}} & 43.1\eb{0.6} \\
 & Countdown & \textbf{51.9\eb{1.3}} & \underline{51.9\eb{1.3}} & \underline{51.3\eb{1.3}} & 51.1\eb{2.0} & 35.7\eb{18.1} & 35.7\eb{18.1} & 51.2\eb{1.7} & \textbf{52.3\eb{2.3}} \\
 & Knights \& Knaves & \textbf{55.1\eb{1.9}} & \underline{55.1\eb{1.9}} & 53.7\eb{0.7} & 54.4\eb{1.1} & 52.9\eb{3.2} & 52.9\eb{3.2} & \underline{54.9\eb{1.6}} & \textbf{56.3\eb{0.7}} \\
 & \emph{Mean} & \textbf{63.8\eb{0.3}} & \underline{63.8\eb{0.3}} & 63.3\eb{0.1} & 63.5\eb{0.8} & 59.0\eb{5.1} & 59.0\eb{5.1} & \underline{63.6\eb{0.2}} & \textbf{64.1\eb{0.4}} \\
\addlinespace[2pt]
Qwen2.5-7B & MATH500 & \textbf{83.7\eb{0.7}} & \textbf{83.7\eb{0.7}} & 83.2\eb{0.3} & 83.3\eb{0.1} & 83.2\eb{1.1} & 83.2\eb{1.1} & \underline{83.4\eb{0.3}} & \underline{83.5\eb{0.6}} \\
 & GSM8K & \textbf{94.8\eb{0.0}} & \textbf{94.8\eb{0.0}} & \underline{94.5\eb{0.2}} & \underline{94.7\eb{0.1}} & 94.1\eb{0.3} & 94.1\eb{0.3} & 94.4\eb{0.1} & 94.5\eb{0.1} \\
 & OlympiadBench & \underline{53.0\eb{0.3}} & \textbf{53.0\eb{0.3}} & \textbf{53.3\eb{0.6}} & \underline{52.7\eb{0.6}} & 50.7\eb{1.5} & 50.7\eb{1.5} & 52.4\eb{0.4} & 52.3\eb{0.9} \\
 & Countdown & \textbf{56.3\eb{1.5}} & \textbf{56.3\eb{1.5}} & \underline{55.2\eb{1.0}} & \underline{55.0\eb{0.9}} & 51.4\eb{2.2} & 51.4\eb{2.2} & 54.0\eb{4.1} & 53.7\eb{2.2} \\
 & Knights \& Knaves & 59.5\eb{1.1} & 59.5\eb{1.1} & \textbf{60.1\eb{1.0}} & 59.2\eb{0.3} & \underline{59.8\eb{2.2}} & \underline{59.8\eb{2.2}} & 59.6\eb{1.5} & \textbf{60.7\eb{1.4}} \\
 & \emph{Mean} & \textbf{69.5\eb{0.2}} & \textbf{69.5\eb{0.2}} & \underline{69.2\eb{0.1}} & \underline{69.0\eb{0.2}} & 67.8\eb{1.2} & 67.8\eb{1.2} & 68.7\eb{1.0} & \underline{69.0\eb{0.4}} \\
\addlinespace[2pt]
Llama-3.1-8B & MATH500 & \underline{66.2\eb{1.0}} & 66.2\eb{1.0} & \textbf{66.8\eb{0.3}} & \underline{66.8\eb{0.6}} & 60.3\eb{1.0} & 60.3\eb{1.0} & \textbf{66.8\eb{0.9}} & \textbf{67.4\eb{0.7}} \\
 & GSM8K & \textbf{92.2\eb{0.4}} & \textbf{92.2\eb{0.4}} & 91.7\eb{0.1} & \underline{92.1\eb{0.3}} & 89.3\eb{0.8} & 89.3\eb{0.8} & \underline{91.8\eb{0.2}} & 91.8\eb{0.1} \\
 & OlympiadBench & 32.9\eb{1.1} & 32.9\eb{1.1} & \underline{33.8\eb{0.5}} & \underline{33.7\eb{0.5}} & 30.0\eb{0.4} & 30.0\eb{0.4} & \textbf{35.0\eb{1.0}} & \textbf{34.4\eb{1.1}} \\
 & Countdown & 46.8\eb{2.1} & 46.8\eb{2.1} & \underline{50.0\eb{1.6}} & \underline{47.2\eb{3.9}} & 38.4\eb{15.8} & 38.4\eb{15.8} & \textbf{54.0\eb{3.1}} & \textbf{51.3\eb{2.4}} \\
 & Knights \& Knaves & 39.9\eb{0.3} & \underline{39.9\eb{0.3}} & \textbf{41.7\eb{1.2}} & 39.1\eb{1.0} & 39.6\eb{5.1} & 39.6\eb{5.1} & \underline{41.5\eb{2.5}} & \textbf{40.3\eb{1.3}} \\
 & \emph{Mean} & 55.6\eb{0.3} & 55.6\eb{0.3} & \underline{56.8\eb{0.5}} & \underline{55.8\eb{0.6}} & 51.5\eb{3.2} & 51.5\eb{3.2} & \textbf{57.8\eb{1.5}} & \textbf{57.0\eb{0.7}} \\
\bottomrule
\end{tabular}

  \end{center}
\end{table}

\begin{table}[!htbp]
  \caption{Per-benchmark accuracy (\%) under confidence-filtered voting, all four methods, $K \in \{2,4\}$; fixed $160$-vote pools; mean\,$\pm$\,std over three seeds. Bold: best method per row and $K$; underlined: second best.}
  \label{tab:lik_perbench}
  \begin{center}
  \scriptsize
  \setlength{\tabcolsep}{2.5pt}
  \begin{tabular}{llcccccccc}
\toprule
 & & \multicolumn{2}{c}{Base} & \multicolumn{2}{c}{\singlem{}} & \multicolumn{2}{c}{\single{}} & \multicolumn{2}{c}{Thicket} \\
\cmidrule(lr){3-4} \cmidrule(lr){5-6} \cmidrule(lr){7-8} \cmidrule(lr){9-10}
Model & Benchmark ($K{=}$) & 2 & 4 & 2 & 4 & 2 & 4 & 2 & 4 \\
\midrule
Qwen2.5-1.5B & MATH500 & \underline{69.7\eb{0.5}} & 69.5\eb{0.6} & 69.6\eb{1.0} & \underline{71.4\eb{1.0}} & 68.1\eb{1.3} & 67.1\eb{0.7} & \textbf{70.9\eb{0.4}} & \textbf{71.9\eb{0.8}} \\
 & GSM8K & \textbf{86.4\eb{0.3}} & \textbf{86.4\eb{0.2}} & 84.1\eb{0.3} & \underline{85.0\eb{0.3}} & 82.4\eb{1.1} & 82.4\eb{1.2} & \underline{85.0\eb{0.4}} & \underline{85.0\eb{0.2}} \\
 & OlympiadBench & \textbf{35.9\eb{0.3}} & 35.4\eb{0.5} & \underline{35.4\eb{1.3}} & \underline{36.2\eb{1.7}} & 31.9\eb{1.3} & 32.5\eb{0.9} & \underline{35.4\eb{1.1}} & \textbf{36.6\eb{0.6}} \\
 & Countdown & 38.0\eb{2.8} & 38.2\eb{2.7} & \underline{51.3\eb{3.2}} & \textbf{54.7\eb{8.0}} & 41.7\eb{10.9} & 42.4\eb{10.5} & \textbf{53.8\eb{9.8}} & \underline{47.8\eb{3.2}} \\
 & Knights \& Knaves & 28.5\eb{0.6} & 29.4\eb{0.6} & \underline{34.9\eb{2.3}} & \underline{33.0\eb{1.3}} & 34.0\eb{3.0} & \textbf{33.2\eb{1.5}} & \textbf{36.2\eb{1.5}} & 32.2\eb{0.4} \\
 & \emph{Mean} & 51.7\eb{0.6} & 51.8\eb{0.4} & \underline{55.0\eb{0.0}} & \textbf{56.1\eb{2.0}} & 51.6\eb{1.7} & 51.5\eb{1.8} & \textbf{56.3\eb{1.7}} & \underline{54.7\eb{0.9}} \\
\addlinespace[2pt]
Qwen2.5-3B & MATH500 & \textbf{78.9\eb{0.4}} & \textbf{78.8\eb{0.4}} & 77.2\eb{0.3} & 76.6\eb{0.7} & 74.4\eb{1.5} & 74.3\eb{2.0} & \underline{77.9\eb{0.3}} & \underline{78.7\eb{1.1}} \\
 & GSM8K & \textbf{91.6\eb{0.5}} & \textbf{91.6\eb{0.2}} & 90.9\eb{0.3} & 91.1\eb{0.6} & 90.0\eb{0.3} & 89.8\eb{0.5} & \underline{91.1\eb{0.6}} & \underline{91.5\eb{0.2}} \\
 & OlympiadBench & \textbf{45.7\eb{1.3}} & \textbf{45.6\eb{0.9}} & 44.3\eb{0.7} & 44.4\eb{1.4} & 42.3\eb{1.9} & 42.3\eb{2.4} & \underline{45.3\eb{1.4}} & \underline{45.3\eb{0.6}} \\
 & Countdown & \textbf{51.7\eb{1.5}} & \underline{51.0\eb{1.5}} & 46.2\eb{4.5} & 50.8\eb{3.5} & 35.7\eb{18.1} & 35.7\eb{18.1} & \underline{47.6\eb{6.1}} & \textbf{52.6\eb{2.6}} \\
 & Knights \& Knaves & \underline{54.2\eb{3.4}} & 54.4\eb{3.0} & \textbf{54.6\eb{2.8}} & \textbf{56.0\eb{2.2}} & 53.2\eb{5.0} & 52.5\eb{5.2} & 53.7\eb{1.3} & \underline{55.3\eb{3.7}} \\
 & \emph{Mean} & \textbf{64.4\eb{0.4}} & \underline{64.3\eb{0.1}} & 62.6\eb{1.3} & 63.8\eb{1.6} & 59.1\eb{5.3} & 58.9\eb{5.5} & \underline{63.1\eb{1.7}} & \textbf{64.7\eb{0.4}} \\
\addlinespace[2pt]
Qwen2.5-7B & MATH500 & \textbf{85.0\eb{0.3}} & \textbf{84.8\eb{0.2}} & 84.3\eb{0.8} & 83.8\eb{0.7} & 83.2\eb{1.2} & 83.4\eb{1.2} & \underline{84.4\eb{0.7}} & \underline{84.3\eb{1.1}} \\
 & GSM8K & \textbf{94.8\eb{0.0}} & \textbf{94.8\eb{0.0}} & \underline{94.4\eb{0.1}} & 94.3\eb{0.3} & 94.0\eb{0.3} & 94.1\eb{0.4} & 94.3\eb{0.1} & \underline{94.5\eb{0.1}} \\
 & OlympiadBench & \textbf{53.3\eb{0.9}} & \textbf{53.9\eb{0.4}} & \underline{52.3\eb{1.5}} & \underline{53.4\eb{1.2}} & 50.4\eb{0.9} & 50.9\eb{1.3} & \underline{52.3\eb{0.5}} & 52.3\eb{0.1} \\
 & Countdown & \textbf{56.3\eb{1.5}} & \textbf{56.3\eb{1.5}} & 51.1\eb{3.0} & 49.7\eb{3.5} & \underline{51.4\eb{2.2}} & 51.4\eb{2.2} & 50.1\eb{5.1} & \underline{52.2\eb{1.0}} \\
 & Knights \& Knaves & 60.1\eb{1.0} & \underline{60.6\eb{1.9}} & \textbf{61.9\eb{3.5}} & 59.2\eb{3.0} & 58.8\eb{1.8} & 58.9\eb{1.9} & \underline{61.1\eb{3.4}} & \textbf{61.6\eb{1.5}} \\
 & \emph{Mean} & \textbf{69.9\eb{0.2}} & \textbf{70.1\eb{0.3}} & \underline{68.8\eb{1.6}} & 68.1\eb{1.7} & 67.6\eb{1.1} & 67.8\eb{1.2} & 68.5\eb{1.0} & \underline{69.0\eb{0.1}} \\
\addlinespace[2pt]
Llama-3.1-8B & MATH500 & \textbf{67.3\eb{0.8}} & \underline{67.3\eb{0.2}} & 64.2\eb{2.7} & \textbf{67.5\eb{0.5}} & 61.4\eb{0.4} & 61.2\eb{1.0} & \underline{66.6\eb{3.2}} & 67.0\eb{1.5} \\
 & GSM8K & \textbf{92.6\eb{0.3}} & \textbf{92.6\eb{0.3}} & 90.7\eb{1.0} & 91.1\eb{0.8} & 89.7\eb{0.8} & 89.8\eb{0.9} & \underline{90.9\eb{1.0}} & \underline{91.3\eb{0.5}} \\
 & OlympiadBench & \underline{33.2\eb{0.5}} & 33.5\eb{0.4} & 32.4\eb{3.5} & \underline{36.0\eb{0.8}} & 30.5\eb{0.4} & 30.2\eb{0.7} & \textbf{34.2\eb{2.0}} & \textbf{36.3\eb{0.8}} \\
 & Countdown & \textbf{48.0\eb{0.6}} & 48.1\eb{0.8} & \underline{43.7\eb{3.9}} & \textbf{58.4\eb{7.5}} & 37.9\eb{15.4} & 38.1\eb{15.6} & 30.8\eb{17.2} & \underline{48.8\eb{10.2}} \\
 & Knights \& Knaves & 40.0\eb{2.3} & 40.5\eb{0.8} & \underline{40.7\eb{2.9}} & \textbf{42.7\eb{0.6}} & 38.8\eb{5.4} & 38.1\eb{5.7} & \textbf{44.9\eb{0.5}} & \underline{42.1\eb{2.2}} \\
 & \emph{Mean} & \textbf{56.2\eb{0.7}} & 56.4\eb{0.0} & \underline{54.3\eb{1.4}} & \textbf{59.1\eb{1.3}} & 51.6\eb{3.2} & 51.5\eb{3.3} & 53.5\eb{4.4} & \underline{57.1\eb{1.2}} \\
\bottomrule
\end{tabular}

  \end{center}
\end{table}

\begin{table}[!htbp]
  \caption{Per-benchmark accuracy (\%) under confidence-filtered voting, $K \in \{8,16\}$ (format of Table~\ref{tab:lik_perbench}).}
  \label{tab:lik_perbench_b}
  \begin{center}
  \scriptsize
  \setlength{\tabcolsep}{2.5pt}
  \begin{tabular}{llcccccccc}
\toprule
 & & \multicolumn{2}{c}{Base} & \multicolumn{2}{c}{\singlem{}} & \multicolumn{2}{c}{\single{}} & \multicolumn{2}{c}{Thicket} \\
\cmidrule(lr){3-4} \cmidrule(lr){5-6} \cmidrule(lr){7-8} \cmidrule(lr){9-10}
Model & Benchmark ($K{=}$) & 8 & 16 & 8 & 16 & 8 & 16 & 8 & 16 \\
\midrule
Qwen2.5-1.5B & MATH500 & 69.9\eb{0.6} & 69.5\eb{0.4} & \underline{70.9\eb{0.2}} & \textbf{70.8\eb{0.7}} & 67.3\eb{1.0} & 67.2\eb{0.7} & \textbf{71.0\eb{0.4}} & \underline{70.0\eb{0.7}} \\
 & GSM8K & \textbf{86.1\eb{0.2}} & \textbf{86.3\eb{0.4}} & \underline{85.8\eb{0.2}} & \textbf{86.3\eb{0.2}} & 82.4\eb{1.1} & \underline{82.4\eb{1.1}} & 85.3\eb{0.5} & \textbf{86.3\eb{0.6}} \\
 & OlympiadBench & \underline{35.8\eb{0.4}} & \underline{34.7\eb{0.8}} & 35.2\eb{0.6} & 34.5\eb{0.7} & 32.6\eb{1.0} & 32.8\eb{1.4} & \textbf{36.6\eb{1.0}} & \textbf{35.3\eb{0.2}} \\
 & Countdown & 38.8\eb{2.1} & 38.1\eb{3.0} & \underline{47.6\eb{2.4}} & \underline{45.0\eb{3.5}} & 42.4\eb{10.5} & 42.4\eb{10.5} & \textbf{53.6\eb{0.3}} & \textbf{49.6\eb{0.6}} \\
 & Knights \& Knaves & 29.4\eb{1.1} & 29.4\eb{0.8} & \underline{31.6\eb{0.6}} & 28.6\eb{2.6} & \textbf{32.8\eb{2.6}} & \textbf{34.4\eb{2.1}} & 31.1\eb{1.6} & \underline{32.6\eb{1.7}} \\
 & \emph{Mean} & 52.0\eb{0.4} & 51.6\eb{0.3} & \underline{54.2\eb{0.4}} & \underline{53.1\eb{0.7}} & 51.5\eb{1.7} & 51.8\eb{1.6} & \textbf{55.5\eb{0.5}} & \textbf{54.8\eb{0.2}} \\
\addlinespace[2pt]
Qwen2.5-3B & MATH500 & \textbf{78.4\eb{0.8}} & \textbf{78.5\eb{1.6}} & 77.5\eb{0.7} & \underline{78.0\eb{0.6}} & 74.5\eb{1.7} & 74.6\eb{2.0} & \underline{77.9\eb{1.4}} & 77.8\eb{1.0} \\
 & GSM8K & \textbf{91.6\eb{0.1}} & \underline{91.4\eb{0.2}} & \textbf{91.6\eb{0.2}} & \textbf{91.6\eb{0.3}} & 89.7\eb{0.5} & 89.8\eb{0.6} & \underline{91.5\eb{0.2}} & \textbf{91.6\eb{0.2}} \\
 & OlympiadBench & \textbf{46.4\eb{0.6}} & \textbf{45.5\eb{0.8}} & 45.7\eb{0.8} & \underline{45.4\eb{0.6}} & 42.6\eb{2.1} & 42.4\eb{2.0} & \underline{46.0\eb{0.4}} & 44.6\eb{0.4} \\
 & Countdown & \textbf{52.1\eb{0.8}} & \underline{51.5\eb{1.9}} & \underline{51.3\eb{1.3}} & 50.6\eb{2.6} & 35.7\eb{18.1} & 35.7\eb{18.1} & 51.2\eb{1.7} & \textbf{52.3\eb{2.3}} \\
 & Knights \& Knaves & \textbf{54.2\eb{3.5}} & \underline{54.0\eb{3.7}} & \underline{53.9\eb{0.7}} & 53.7\eb{1.3} & 53.2\eb{5.2} & 53.3\eb{4.8} & \textbf{54.2\eb{1.6}} & \textbf{56.3\eb{0.7}} \\
 & \emph{Mean} & \textbf{64.6\eb{0.4}} & \underline{64.2\eb{0.4}} & 64.0\eb{0.4} & 63.9\eb{0.8} & 59.2\eb{5.4} & 59.2\eb{5.5} & \underline{64.2\eb{0.2}} & \textbf{64.5\eb{0.4}} \\
\addlinespace[2pt]
Qwen2.5-7B & MATH500 & \textbf{84.9\eb{0.7}} & \textbf{84.5\eb{0.8}} & \underline{84.7\eb{0.7}} & 83.7\eb{0.8} & 83.5\eb{0.7} & 83.3\eb{1.1} & 84.5\eb{0.7} & \underline{84.4\eb{0.7}} \\
 & GSM8K & \textbf{94.8\eb{0.0}} & \textbf{94.8\eb{0.0}} & \underline{94.5\eb{0.2}} & \underline{94.7\eb{0.2}} & 94.1\eb{0.5} & 94.1\eb{0.5} & 94.4\eb{0.1} & 94.4\eb{0.2} \\
 & OlympiadBench & \textbf{53.3\eb{0.5}} & \textbf{53.7\eb{0.7}} & \textbf{53.3\eb{0.3}} & \underline{53.3\eb{0.1}} & 51.1\eb{1.2} & 50.3\eb{1.2} & \underline{52.4\eb{0.8}} & 52.7\eb{1.1} \\
 & Countdown & \textbf{56.3\eb{1.5}} & \textbf{56.3\eb{1.5}} & \underline{55.2\eb{1.0}} & \underline{55.0\eb{0.9}} & 51.4\eb{2.2} & 51.4\eb{2.2} & 54.0\eb{4.1} & 53.7\eb{2.2} \\
 & Knights \& Knaves & \underline{60.5\eb{2.0}} & \underline{59.9\eb{0.9}} & 60.1\eb{1.6} & 59.1\eb{1.6} & 59.1\eb{1.6} & 59.3\eb{1.7} & \textbf{61.4\eb{1.5}} & \textbf{61.6\eb{2.9}} \\
 & \emph{Mean} & \textbf{70.0\eb{0.6}} & \textbf{69.8\eb{0.2}} & \underline{69.6\eb{0.6}} & 69.2\eb{0.3} & 67.8\eb{1.0} & 67.7\eb{1.0} & 69.3\eb{1.0} & \underline{69.4\eb{0.4}} \\
\addlinespace[2pt]
Llama-3.1-8B & MATH500 & \underline{67.6\eb{0.2}} & 67.3\eb{0.7} & 67.3\eb{1.8} & \underline{67.5\eb{0.2}} & 61.3\eb{0.5} & 61.1\eb{0.3} & \textbf{67.8\eb{0.9}} & \textbf{68.6\eb{0.6}} \\
 & GSM8K & \textbf{92.6\eb{0.4}} & \textbf{92.4\eb{0.2}} & \underline{91.7\eb{0.4}} & \underline{92.3\eb{0.1}} & 89.6\eb{0.8} & 89.7\eb{0.6} & 91.6\eb{0.4} & 91.8\eb{0.3} \\
 & OlympiadBench & 33.4\eb{0.3} & 33.3\eb{0.8} & \underline{34.5\eb{0.8}} & \underline{33.7\eb{0.6}} & 30.6\eb{0.7} & 30.5\eb{0.3} & \textbf{36.3\eb{1.1}} & \textbf{35.6\eb{0.3}} \\
 & Countdown & 47.9\eb{1.0} & \underline{47.0\eb{2.0}} & \underline{49.7\eb{0.8}} & 46.6\eb{4.9} & 37.4\eb{15.0} & 37.7\eb{15.2} & \textbf{55.0\eb{1.7}} & \textbf{51.5\eb{3.7}} \\
 & Knights \& Knaves & 38.8\eb{2.6} & \underline{39.8\eb{0.3}} & \underline{41.5\eb{1.3}} & 39.6\eb{1.1} & 39.4\eb{5.4} & 38.0\eb{5.4} & \textbf{41.6\eb{2.3}} & \textbf{40.5\eb{0.7}} \\
 & \emph{Mean} & 56.0\eb{0.5} & \underline{55.9\eb{0.5}} & \underline{56.9\eb{0.9}} & \underline{55.9\eb{1.1}} & 51.6\eb{3.1} & 51.4\eb{3.1} & \textbf{58.4\eb{1.1}} & \textbf{57.6\eb{0.7}} \\
\bottomrule
\end{tabular}

  \end{center}
\end{table}

\begin{table}[!htbp]
  \caption{\textbf{Thicket accuracy under confidence-filtered voting} (each voter keeps its top-$q$ completions by mean log-likelihood, $q$ chosen by per-benchmark five-fold CV; fixed $160$-vote pools; mean\,$\pm$\,std over three seeds). Mean column: macro over the five benchmarks. Bold: the displayed value improves on the same setting's plain majority vote.}
  \label{tab:lik}
  \begin{center}
  \footnotesize
  \setlength{\tabcolsep}{4pt}
  \begin{tabular}{llcccccc}
\toprule
Model & $K$ & MATH500 & GSM8K & OlympiadBench & Countdown & Knights \& Knaves & \emph{Mean} \\
\midrule
Qwen2.5-1.5B & 2 & \textbf{70.9\eb{0.4}} & \textbf{85.0\eb{0.4}} & \textbf{35.4\eb{1.1}} & 53.8\eb{9.8} & \textbf{36.2\eb{1.5}} & \textbf{56.3\eb{1.7}} \\
 & 4 & \textbf{71.9\eb{0.8}} & 85.0\eb{0.2} & \textbf{36.6\eb{0.6}} & 47.8\eb{3.2} & \textbf{32.2\eb{0.4}} & \textbf{54.7\eb{0.9}} \\
 & 8 & \textbf{71.0\eb{0.4}} & 85.3\eb{0.5} & \textbf{36.6\eb{1.0}} & 53.6\eb{0.3} & 31.1\eb{1.6} & \textbf{55.5\eb{0.5}} \\
 & 16 & \textbf{70.0\eb{0.7}} & \textbf{86.3\eb{0.6}} & \textbf{35.3\eb{0.2}} & 49.6\eb{0.6} & 32.6\eb{1.7} & \textbf{54.8\eb{0.2}} \\
\addlinespace[2pt]
Qwen2.5-3B & 2 & \textbf{77.9\eb{0.3}} & \textbf{91.1\eb{0.6}} & \textbf{45.3\eb{1.4}} & 47.6\eb{6.1} & 53.7\eb{1.3} & \textbf{63.1\eb{1.7}} \\
 & 4 & \textbf{78.7\eb{1.1}} & \textbf{91.5\eb{0.2}} & \textbf{45.3\eb{0.6}} & 52.6\eb{2.6} & 55.3\eb{3.7} & \textbf{64.7\eb{0.4}} \\
 & 8 & \textbf{77.9\eb{1.4}} & 91.5\eb{0.2} & \textbf{46.0\eb{0.4}} & 51.2\eb{1.7} & 54.2\eb{1.6} & \textbf{64.2\eb{0.2}} \\
 & 16 & \textbf{77.8\eb{1.0}} & 91.6\eb{0.2} & \textbf{44.6\eb{0.4}} & 52.3\eb{2.3} & 56.3\eb{0.7} & \textbf{64.5\eb{0.4}} \\
\addlinespace[2pt]
Qwen2.5-7B & 2 & \textbf{84.4\eb{0.7}} & 94.3\eb{0.1} & 52.3\eb{0.5} & 50.1\eb{5.1} & 61.1\eb{3.4} & 68.5\eb{1.0} \\
 & 4 & \textbf{84.3\eb{1.1}} & 94.5\eb{0.1} & 52.3\eb{0.1} & 52.2\eb{1.0} & 61.6\eb{1.5} & \textbf{69.0\eb{0.1}} \\
 & 8 & \textbf{84.5\eb{0.7}} & 94.4\eb{0.1} & 52.4\eb{0.8} & 54.0\eb{4.1} & \textbf{61.4\eb{1.5}} & \textbf{69.3\eb{1.0}} \\
 & 16 & \textbf{84.4\eb{0.7}} & 94.4\eb{0.2} & \textbf{52.7\eb{1.1}} & 53.7\eb{2.2} & \textbf{61.6\eb{2.9}} & \textbf{69.4\eb{0.4}} \\
\addlinespace[2pt]
Llama-3.1-8B & 2 & \textbf{66.6\eb{3.2}} & \textbf{90.9\eb{1.0}} & \textbf{34.2\eb{2.0}} & 30.8\eb{17.2} & \textbf{44.9\eb{0.5}} & \textbf{53.5\eb{4.4}} \\
 & 4 & \textbf{67.0\eb{1.5}} & \textbf{91.3\eb{0.5}} & \textbf{36.3\eb{0.8}} & 48.8\eb{10.2} & 42.1\eb{2.2} & \textbf{57.1\eb{1.2}} \\
 & 8 & \textbf{67.8\eb{0.9}} & 91.6\eb{0.4} & \textbf{36.3\eb{1.1}} & \textbf{55.0\eb{1.7}} & \textbf{41.6\eb{2.3}} & \textbf{58.4\eb{1.1}} \\
 & 16 & \textbf{68.6\eb{0.6}} & 91.8\eb{0.3} & \textbf{35.6\eb{0.3}} & \textbf{51.5\eb{3.7}} & \textbf{40.5\eb{0.7}} & \textbf{57.6\eb{0.7}} \\
\bottomrule
\end{tabular}

  \end{center}
\end{table}

\begin{table}[!htbp]
  \caption{Base-model accuracy under confidence-filtered voting (format of Table~\ref{tab:lik}; three seeds).}
  \label{tab:lik_base}
  \begin{center}
  \footnotesize
  \setlength{\tabcolsep}{4pt}
  \begin{tabular}{llcccccc}
\toprule
Model & $K$ & MATH500 & GSM8K & OlympiadBench & Countdown & Knights \& Knaves & \emph{Mean} \\
\midrule
Qwen2.5-1.5B & 2 & \textbf{69.7\eb{0.5}} & 86.4\eb{0.3} & \textbf{35.9\eb{0.3}} & 38.0\eb{2.8} & 28.5\eb{0.6} & \textbf{51.7\eb{0.6}} \\
 & 4 & \textbf{69.5\eb{0.6}} & 86.4\eb{0.2} & \textbf{35.4\eb{0.5}} & 38.2\eb{2.7} & 29.4\eb{0.6} & \textbf{51.8\eb{0.4}} \\
 & 8 & \textbf{69.9\eb{0.6}} & 86.1\eb{0.2} & \textbf{35.8\eb{0.4}} & 38.8\eb{2.1} & 29.4\eb{1.1} & \textbf{52.0\eb{0.4}} \\
 & 16 & \textbf{69.5\eb{0.4}} & 86.3\eb{0.4} & \textbf{34.7\eb{0.8}} & 38.1\eb{3.0} & 29.4\eb{0.8} & \textbf{51.6\eb{0.3}} \\
\addlinespace[2pt]
Qwen2.5-3B & 2 & \textbf{78.9\eb{0.4}} & 91.6\eb{0.5} & \textbf{45.7\eb{1.3}} & 51.7\eb{1.5} & 54.2\eb{3.4} & \textbf{64.4\eb{0.4}} \\
 & 4 & \textbf{78.8\eb{0.4}} & 91.6\eb{0.2} & \textbf{45.6\eb{0.9}} & 51.0\eb{1.5} & 54.4\eb{3.0} & \textbf{64.3\eb{0.1}} \\
 & 8 & \textbf{78.4\eb{0.8}} & 91.6\eb{0.1} & \textbf{46.4\eb{0.6}} & \textbf{52.1\eb{0.8}} & 54.2\eb{3.5} & \textbf{64.6\eb{0.4}} \\
 & 16 & \textbf{78.5\eb{1.6}} & 91.4\eb{0.2} & \textbf{45.5\eb{0.8}} & 51.5\eb{1.9} & 54.0\eb{3.7} & \textbf{64.2\eb{0.4}} \\
\addlinespace[2pt]
Qwen2.5-7B & 2 & \textbf{85.0\eb{0.3}} & 94.8\eb{0.0} & \textbf{53.3\eb{0.9}} & 56.3\eb{1.5} & \textbf{60.1\eb{1.0}} & \textbf{69.9\eb{0.2}} \\
 & 4 & \textbf{84.8\eb{0.2}} & 94.8\eb{0.0} & \textbf{53.9\eb{0.4}} & 56.3\eb{1.5} & \textbf{60.6\eb{1.9}} & \textbf{70.1\eb{0.3}} \\
 & 8 & \textbf{84.9\eb{0.7}} & 94.8\eb{0.0} & \textbf{53.3\eb{0.5}} & 56.3\eb{1.5} & \textbf{60.5\eb{2.0}} & \textbf{70.0\eb{0.6}} \\
 & 16 & \textbf{84.5\eb{0.8}} & 94.8\eb{0.0} & \textbf{53.7\eb{0.7}} & 56.3\eb{1.5} & \textbf{59.9\eb{0.9}} & \textbf{69.8\eb{0.2}} \\
\addlinespace[2pt]
Llama-3.1-8B & 2 & \textbf{67.3\eb{0.8}} & \textbf{92.6\eb{0.3}} & \textbf{33.2\eb{0.5}} & \textbf{48.0\eb{0.6}} & \textbf{40.0\eb{2.3}} & \textbf{56.2\eb{0.7}} \\
 & 4 & \textbf{67.3\eb{0.2}} & \textbf{92.6\eb{0.3}} & \textbf{33.5\eb{0.4}} & \textbf{48.1\eb{0.8}} & \textbf{40.5\eb{0.8}} & \textbf{56.4\eb{0.0}} \\
 & 8 & \textbf{67.6\eb{0.2}} & \textbf{92.6\eb{0.4}} & \textbf{33.4\eb{0.3}} & \textbf{47.9\eb{1.0}} & 38.8\eb{2.6} & \textbf{56.0\eb{0.5}} \\
 & 16 & \textbf{67.3\eb{0.7}} & \textbf{92.4\eb{0.2}} & \textbf{33.3\eb{0.8}} & \textbf{47.0\eb{2.0}} & 39.8\eb{0.3} & \textbf{55.9\eb{0.5}} \\
\bottomrule
\end{tabular}

  \end{center}
\end{table}

\begin{table}[!htbp]
  \caption{\single{} accuracy under confidence-filtered voting (format of Table~\ref{tab:lik}; three seeds).}
  \label{tab:lik_single}
  \begin{center}
  \footnotesize
  \setlength{\tabcolsep}{4pt}
  \begin{tabular}{llcccccc}
\toprule
Model & $K$ & MATH500 & GSM8K & OlympiadBench & Countdown & Knights \& Knaves & \emph{Mean} \\
\midrule
Qwen2.5-1.5B & 2 & \textbf{68.1\eb{1.3}} & 82.4\eb{1.1} & 31.9\eb{1.3} & 41.7\eb{10.9} & \textbf{34.0\eb{3.0}} & 51.6\eb{1.7} \\
 & 4 & 67.1\eb{0.7} & 82.4\eb{1.2} & \textbf{32.5\eb{0.9}} & 42.4\eb{10.5} & 33.2\eb{1.5} & 51.5\eb{1.8} \\
 & 8 & 67.3\eb{1.0} & 82.4\eb{1.1} & \textbf{32.6\eb{1.0}} & 42.4\eb{10.5} & 32.8\eb{2.6} & 51.5\eb{1.7} \\
 & 16 & 67.2\eb{0.7} & 82.4\eb{1.1} & \textbf{32.8\eb{1.4}} & 42.4\eb{10.5} & \textbf{34.4\eb{2.1}} & \textbf{51.8\eb{1.6}} \\
\addlinespace[2pt]
Qwen2.5-3B & 2 & 74.4\eb{1.5} & \textbf{90.0\eb{0.3}} & \textbf{42.3\eb{1.9}} & 35.7\eb{18.1} & \textbf{53.2\eb{5.0}} & \textbf{59.1\eb{5.3}} \\
 & 4 & 74.3\eb{2.0} & 89.8\eb{0.5} & \textbf{42.3\eb{2.4}} & 35.7\eb{18.1} & 52.5\eb{5.2} & 58.9\eb{5.5} \\
 & 8 & 74.5\eb{1.7} & 89.7\eb{0.5} & \textbf{42.6\eb{2.1}} & 35.7\eb{18.1} & \textbf{53.2\eb{5.2}} & \textbf{59.2\eb{5.4}} \\
 & 16 & 74.6\eb{2.0} & 89.8\eb{0.6} & \textbf{42.4\eb{2.0}} & 35.7\eb{18.1} & \textbf{53.3\eb{4.8}} & \textbf{59.2\eb{5.5}} \\
\addlinespace[2pt]
Qwen2.5-7B & 2 & 83.2\eb{1.2} & 94.0\eb{0.3} & 50.4\eb{0.9} & 51.4\eb{2.2} & 58.8\eb{1.8} & 67.6\eb{1.1} \\
 & 4 & \textbf{83.4\eb{1.2}} & 94.1\eb{0.4} & \textbf{50.9\eb{1.3}} & 51.4\eb{2.2} & 58.9\eb{1.9} & 67.8\eb{1.2} \\
 & 8 & \textbf{83.5\eb{0.7}} & 94.1\eb{0.5} & \textbf{51.1\eb{1.2}} & 51.4\eb{2.2} & 59.1\eb{1.6} & 67.8\eb{1.0} \\
 & 16 & \textbf{83.3\eb{1.1}} & 94.1\eb{0.5} & 50.3\eb{1.2} & 51.4\eb{2.2} & 59.3\eb{1.7} & 67.7\eb{1.0} \\
\addlinespace[2pt]
Llama-3.1-8B & 2 & \textbf{61.4\eb{0.4}} & \textbf{89.7\eb{0.8}} & \textbf{30.5\eb{0.4}} & 37.9\eb{15.4} & 38.8\eb{5.4} & \textbf{51.6\eb{3.2}} \\
 & 4 & \textbf{61.2\eb{1.0}} & \textbf{89.8\eb{0.9}} & \textbf{30.2\eb{0.7}} & 38.1\eb{15.6} & 38.1\eb{5.7} & 51.5\eb{3.3} \\
 & 8 & \textbf{61.3\eb{0.5}} & \textbf{89.6\eb{0.8}} & \textbf{30.6\eb{0.7}} & 37.4\eb{15.0} & 39.4\eb{5.4} & \textbf{51.6\eb{3.1}} \\
 & 16 & \textbf{61.1\eb{0.3}} & \textbf{89.7\eb{0.6}} & \textbf{30.5\eb{0.3}} & 37.7\eb{15.2} & 38.0\eb{5.4} & 51.4\eb{3.1} \\
\bottomrule
\end{tabular}

  \end{center}
\end{table}

\begin{table}[!htbp]
  \caption{\singlem{} accuracy under confidence-filtered voting (format of Table~\ref{tab:lik}; three seeds).}
  \label{tab:lik_singlem}
  \begin{center}
  \footnotesize
  \setlength{\tabcolsep}{4pt}
  \begin{tabular}{llcccccc}
\toprule
Model & $K$ & MATH500 & GSM8K & OlympiadBench & Countdown & Knights \& Knaves & \emph{Mean} \\
\midrule
Qwen2.5-1.5B & 2 & \textbf{69.6\eb{1.0}} & 84.1\eb{0.3} & \textbf{35.4\eb{1.3}} & 51.3\eb{3.2} & 34.9\eb{2.3} & \textbf{55.0\eb{0.0}} \\
 & 4 & \textbf{71.4\eb{1.0}} & 85.0\eb{0.3} & \textbf{36.2\eb{1.7}} & 54.7\eb{8.0} & 33.0\eb{1.3} & \textbf{56.1\eb{2.0}} \\
 & 8 & \textbf{70.9\eb{0.2}} & 85.8\eb{0.2} & \textbf{35.2\eb{0.6}} & 47.6\eb{2.4} & 31.6\eb{0.6} & \textbf{54.2\eb{0.4}} \\
 & 16 & \textbf{70.8\eb{0.7}} & \textbf{86.3\eb{0.2}} & \textbf{34.5\eb{0.7}} & 45.0\eb{3.5} & 28.6\eb{2.6} & \textbf{53.1\eb{0.7}} \\
\addlinespace[2pt]
Qwen2.5-3B & 2 & \textbf{77.2\eb{0.3}} & \textbf{90.9\eb{0.3}} & \textbf{44.3\eb{0.7}} & 46.2\eb{4.5} & 54.6\eb{2.8} & \textbf{62.6\eb{1.3}} \\
 & 4 & \textbf{76.6\eb{0.7}} & 91.1\eb{0.6} & \textbf{44.4\eb{1.4}} & 50.8\eb{3.5} & 56.0\eb{2.2} & \textbf{63.8\eb{1.6}} \\
 & 8 & \textbf{77.5\eb{0.7}} & 91.6\eb{0.2} & \textbf{45.7\eb{0.8}} & 51.3\eb{1.3} & \textbf{53.9\eb{0.7}} & \textbf{64.0\eb{0.4}} \\
 & 16 & \textbf{78.0\eb{0.6}} & 91.6\eb{0.3} & \textbf{45.4\eb{0.6}} & 50.6\eb{2.6} & 53.7\eb{1.3} & \textbf{63.9\eb{0.8}} \\
\addlinespace[2pt]
Qwen2.5-7B & 2 & \textbf{84.3\eb{0.8}} & 94.4\eb{0.1} & 52.3\eb{1.5} & 51.1\eb{3.0} & \textbf{61.9\eb{3.5}} & \textbf{68.8\eb{1.6}} \\
 & 4 & \textbf{83.8\eb{0.7}} & 94.3\eb{0.3} & \textbf{53.4\eb{1.2}} & 49.7\eb{3.5} & 59.2\eb{3.0} & \textbf{68.1\eb{1.7}} \\
 & 8 & \textbf{84.7\eb{0.7}} & 94.5\eb{0.2} & 53.3\eb{0.3} & 55.2\eb{1.0} & 60.1\eb{1.6} & \textbf{69.6\eb{0.6}} \\
 & 16 & \textbf{83.7\eb{0.8}} & 94.7\eb{0.2} & \textbf{53.3\eb{0.1}} & 55.0\eb{0.9} & 59.1\eb{1.6} & \textbf{69.2\eb{0.3}} \\
\addlinespace[2pt]
Llama-3.1-8B & 2 & \textbf{64.2\eb{2.7}} & 90.7\eb{1.0} & \textbf{32.4\eb{3.5}} & 43.7\eb{3.9} & 40.7\eb{2.9} & \textbf{54.3\eb{1.4}} \\
 & 4 & \textbf{67.5\eb{0.5}} & \textbf{91.1\eb{0.8}} & \textbf{36.0\eb{0.8}} & \textbf{58.4\eb{7.5}} & 42.7\eb{0.6} & \textbf{59.1\eb{1.3}} \\
 & 8 & \textbf{67.3\eb{1.8}} & 91.7\eb{0.4} & \textbf{34.5\eb{0.8}} & 49.7\eb{0.8} & 41.5\eb{1.3} & \textbf{56.9\eb{0.9}} \\
 & 16 & \textbf{67.5\eb{0.2}} & \textbf{92.3\eb{0.1}} & 33.7\eb{0.6} & 46.6\eb{4.9} & \textbf{39.6\eb{1.1}} & \textbf{55.9\eb{1.1}} \\
\bottomrule
\end{tabular}

  \end{center}
\end{table}

\clearpage

\section{Per-benchmark coverage}
\label{app:cov}

\begin{figure}[!htbp]
    \centering
    \includegraphics[width=0.95\linewidth]{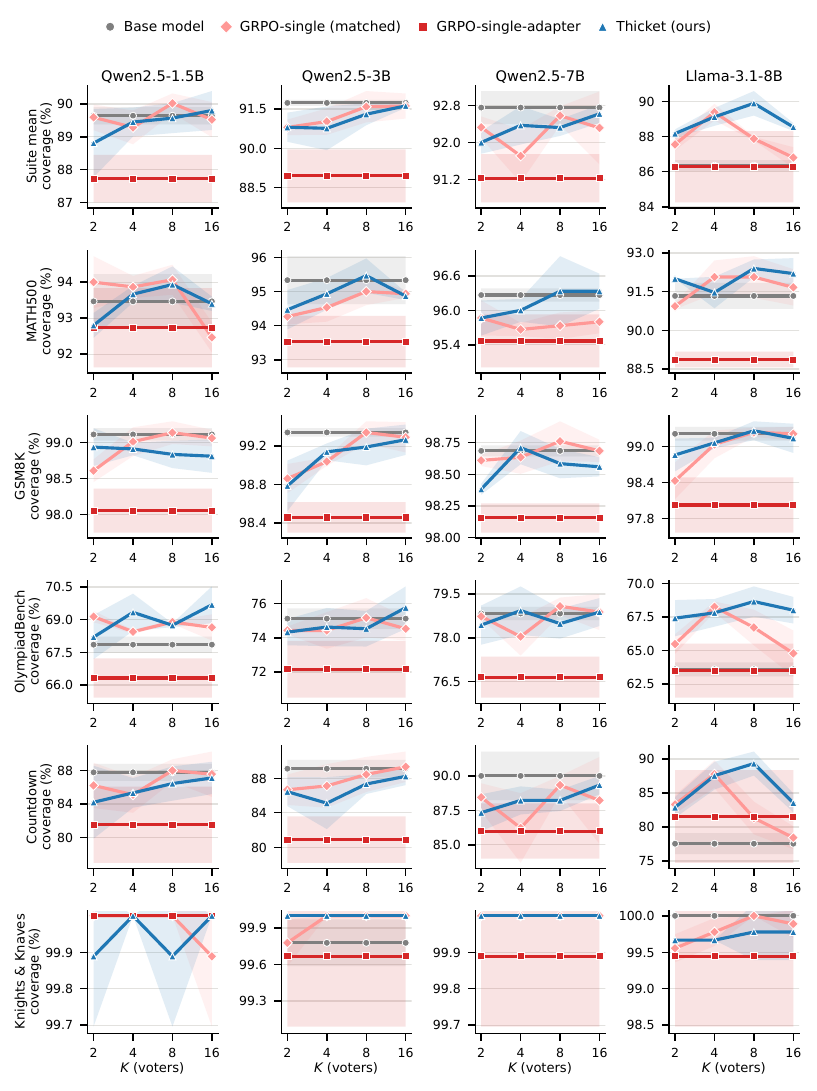}
    \caption{Coverage vs.\ voter count $K$ on the math suite (fixed $160$-completion pools). Lines: seed means; bands: $\pm 1$ std. Base and \single{} are flat by construction. In the Knights~\&~Knaves row, coverage saturates at $100\%$ and coinciding lines hide one another.}
    \label{fig:scaling_cov_math}
\end{figure}

\begin{figure}[!htbp]
    \centering
    \includegraphics[width=0.95\linewidth]{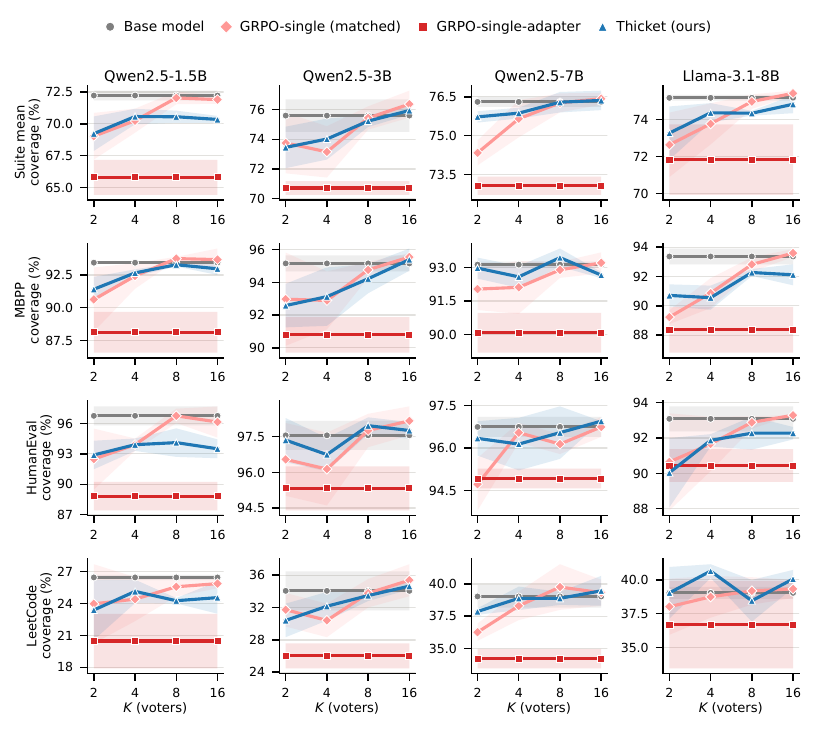}
    \caption{Coverage vs.\ voter count $K$ on the code suite (fixed $160$-completion pools). Lines: seed means; bands: $\pm 1$ std. Base and \single{} are flat by construction.}
    \label{fig:scaling_cov_code}
\end{figure}

Tables~\ref{tab:cov_math_perbench}--\ref{tab:cov_code_perbench_b} break the suite-mean coverage of Table~\ref{tab:cov} down by benchmark. Figures~\ref{fig:scaling_cov_math} and \ref{fig:scaling_cov_code} plot the same numbers as scaling curves with $\pm 1$ std bands.

\begin{table}[!htbp]
  \caption{Per-benchmark coverage (pass@pool \%) on the math suite, $K \in \{2,4\}$; mean\,$\pm$\,std over three seeds. Bold: best method per row and $K$; underlined: second best.}
  \label{tab:cov_math_perbench}
  \begin{center}
  \scriptsize
  \setlength{\tabcolsep}{2.5pt}
  \begin{tabular}{llcccccccc}
\toprule
 & & \multicolumn{2}{c}{Base} & \multicolumn{2}{c}{\singlem{}} & \multicolumn{2}{c}{\single{}} & \multicolumn{2}{c}{Thicket} \\
\cmidrule(lr){3-4} \cmidrule(lr){5-6} \cmidrule(lr){7-8} \cmidrule(lr){9-10}
Model & Benchmark ($K{=}$) & 2 & 4 & 2 & 4 & 2 & 4 & 2 & 4 \\
\midrule
Qwen2.5-1.5B & MATH500 & \underline{93.5\eb{0.8}} & 93.5\eb{0.8} & \textbf{94.0\eb{0.7}} & \textbf{93.9\eb{0.3}} & 92.7\eb{1.1} & 92.7\eb{1.1} & 92.8\eb{0.3} & \underline{93.7\eb{0.2}} \\
 & GSM8K & \textbf{99.1\eb{0.1}} & \textbf{99.1\eb{0.1}} & 98.6\eb{0.2} & \underline{99.0\eb{0.2}} & 98.1\eb{0.3} & 98.1\eb{0.3} & \underline{98.9\eb{0.3}} & 98.9\eb{0.1} \\
 & OlympiadBench & 67.9\eb{0.4} & 67.9\eb{0.4} & \textbf{69.1\eb{0.3}} & \underline{68.4\eb{0.1}} & 66.3\eb{0.9} & 66.3\eb{0.9} & \underline{68.2\eb{1.0}} & \textbf{69.3\eb{0.9}} \\
 & Countdown & \textbf{87.8\eb{1.0}} & \textbf{87.8\eb{1.0}} & \underline{86.2\eb{2.7}} & 85.1\eb{2.1} & 81.6\eb{4.5} & 81.6\eb{4.5} & 84.2\eb{4.3} & \underline{85.3\eb{1.8}} \\
 & Knights \& Knaves & \textbf{100.0\eb{0.0}} & \textbf{100.0\eb{0.0}} & \textbf{100.0\eb{0.0}} & \textbf{100.0\eb{0.0}} & \textbf{100.0\eb{0.0}} & \textbf{100.0\eb{0.0}} & \underline{99.9\eb{0.2}} & \textbf{100.0\eb{0.0}} \\
 & \emph{Mean} & \textbf{89.6\eb{0.3}} & \textbf{89.6\eb{0.3}} & \textbf{89.6\eb{0.4}} & 89.3\eb{0.5} & 87.7\eb{0.7} & 87.7\eb{0.7} & \underline{88.8\eb{1.0}} & \underline{89.5\eb{0.4}} \\
\addlinespace[2pt]
Qwen2.5-3B & MATH500 & \textbf{95.3\eb{0.7}} & \textbf{95.3\eb{0.7}} & 94.3\eb{0.2} & 94.5\eb{0.4} & 93.5\eb{0.8} & 93.5\eb{0.8} & \underline{94.5\eb{0.6}} & \underline{94.9\eb{0.5}} \\
 & GSM8K & \textbf{99.3\eb{0.0}} & \textbf{99.3\eb{0.0}} & \underline{98.9\eb{0.2}} & 99.0\eb{0.0} & 98.5\eb{0.2} & 98.5\eb{0.2} & 98.8\eb{0.3} & \underline{99.1\eb{0.1}} \\
 & OlympiadBench & \textbf{75.1\eb{0.6}} & \textbf{75.1\eb{0.6}} & \underline{74.4\eb{0.2}} & 74.4\eb{1.1} & 72.2\eb{1.7} & 72.2\eb{1.7} & 74.3\eb{0.8} & \underline{74.6\eb{1.1}} \\
 & Countdown & \textbf{89.1\eb{1.0}} & \textbf{89.1\eb{1.0}} & \underline{86.7\eb{1.8}} & \underline{87.1\eb{2.5}} & 80.9\eb{2.7} & 80.9\eb{2.7} & 86.4\eb{1.7} & 85.1\eb{3.0} \\
 & Knights \& Knaves & \underline{99.8\eb{0.2}} & \underline{99.8\eb{0.2}} & \underline{99.8\eb{0.2}} & \textbf{100.0\eb{0.0}} & 99.7\eb{0.6} & 99.7\eb{0.6} & \textbf{100.0\eb{0.0}} & \textbf{100.0\eb{0.0}} \\
 & \emph{Mean} & \textbf{91.7\eb{0.1}} & \textbf{91.7\eb{0.1}} & \underline{90.8\eb{0.3}} & \underline{91.0\eb{0.5}} & 88.9\eb{1.0} & 88.9\eb{1.0} & \underline{90.8\eb{0.6}} & 90.8\eb{0.8} \\
\addlinespace[2pt]
Qwen2.5-7B & MATH500 & \textbf{96.3\eb{0.1}} & \textbf{96.3\eb{0.1}} & \underline{95.9\eb{0.3}} & 95.7\eb{0.1} & 95.5\eb{0.5} & 95.5\eb{0.5} & \underline{95.9\eb{0.3}} & \underline{96.0\eb{0.2}} \\
 & GSM8K & \textbf{98.7\eb{0.0}} & \textbf{98.7\eb{0.0}} & \underline{98.6\eb{0.0}} & \underline{98.6\eb{0.1}} & 98.2\eb{0.1} & 98.2\eb{0.1} & 98.4\eb{0.0} & \textbf{98.7\eb{0.1}} \\
 & OlympiadBench & \textbf{78.8\eb{0.2}} & \underline{78.8\eb{0.2}} & \underline{78.7\eb{0.1}} & 78.0\eb{0.6} & 76.7\eb{0.7} & 76.7\eb{0.7} & 78.4\eb{0.7} & \textbf{78.9\eb{0.8}} \\
 & Countdown & \textbf{90.0\eb{1.8}} & \textbf{90.0\eb{1.8}} & \underline{88.4\eb{1.0}} & 86.2\eb{2.5} & 86.0\eb{2.0} & 86.0\eb{2.0} & 87.3\eb{1.3} & \underline{88.2\eb{1.0}} \\
 & Knights \& Knaves & \textbf{100.0\eb{0.0}} & \textbf{100.0\eb{0.0}} & \textbf{100.0\eb{0.0}} & \textbf{100.0\eb{0.0}} & \underline{99.9\eb{0.2}} & \underline{99.9\eb{0.2}} & \textbf{100.0\eb{0.0}} & \textbf{100.0\eb{0.0}} \\
 & \emph{Mean} & \textbf{92.8\eb{0.4}} & \textbf{92.8\eb{0.4}} & \underline{92.3\eb{0.2}} & 91.7\eb{0.6} & 91.2\eb{0.5} & 91.2\eb{0.5} & 92.0\eb{0.2} & \underline{92.4\eb{0.4}} \\
\addlinespace[2pt]
Llama-3.1-8B & MATH500 & \underline{91.3\eb{0.5}} & 91.3\eb{0.5} & 90.9\eb{0.1} & \textbf{92.1\eb{0.6}} & 88.9\eb{0.3} & 88.9\eb{0.3} & \textbf{92.0\eb{0.0}} & \underline{91.5\eb{0.6}} \\
 & GSM8K & \textbf{99.2\eb{0.1}} & \textbf{99.2\eb{0.1}} & 98.4\eb{0.3} & 99.0\eb{0.2} & 98.0\eb{0.5} & 98.0\eb{0.5} & \underline{98.9\eb{0.3}} & \underline{99.1\eb{0.1}} \\
 & OlympiadBench & 63.6\eb{0.5} & 63.6\eb{0.5} & \underline{65.5\eb{0.7}} & \textbf{68.2\eb{0.4}} & 63.5\eb{2.0} & 63.5\eb{2.0} & \textbf{67.4\eb{1.3}} & \underline{67.8\eb{1.0}} \\
 & Countdown & 77.6\eb{1.5} & 77.6\eb{1.5} & \textbf{83.3\eb{1.3}} & \textbf{87.8\eb{2.0}} & 81.6\eb{6.8} & 81.6\eb{6.8} & \underline{82.9\eb{1.5}} & \underline{87.6\eb{2.0}} \\
 & Knights \& Knaves & \textbf{100.0\eb{0.0}} & \textbf{100.0\eb{0.0}} & 99.6\eb{0.2} & \underline{99.8\eb{0.2}} & 99.4\eb{1.0} & 99.4\eb{1.0} & \underline{99.7\eb{0.0}} & 99.7\eb{0.0} \\
 & \emph{Mean} & 86.3\eb{0.3} & 86.3\eb{0.3} & \underline{87.5\eb{0.3}} & \textbf{89.4\eb{0.5}} & 86.3\eb{2.0} & 86.3\eb{2.0} & \textbf{88.2\eb{0.2}} & \underline{89.1\eb{0.5}} \\
\bottomrule
\end{tabular}

  \end{center}
\end{table}

\begin{table}[!htbp]
  \caption{Per-benchmark coverage (pass@pool \%) on the math suite, $K \in \{8,16\}$ (format of Table~\ref{tab:cov_math_perbench}).}
  \label{tab:cov_math_perbench_b}
  \begin{center}
  \scriptsize
  \setlength{\tabcolsep}{2.5pt}
  \begin{tabular}{llcccccccc}
\toprule
 & & \multicolumn{2}{c}{Base} & \multicolumn{2}{c}{\singlem{}} & \multicolumn{2}{c}{\single{}} & \multicolumn{2}{c}{Thicket} \\
\cmidrule(lr){3-4} \cmidrule(lr){5-6} \cmidrule(lr){7-8} \cmidrule(lr){9-10}
Model & Benchmark ($K{=}$) & 8 & 16 & 8 & 16 & 8 & 16 & 8 & 16 \\
\midrule
Qwen2.5-1.5B & MATH500 & 93.5\eb{0.8} & \textbf{93.5\eb{0.8}} & \textbf{94.1\eb{0.4}} & 92.5\eb{0.4} & 92.7\eb{1.1} & 92.7\eb{1.1} & \underline{93.9\eb{0.5}} & \underline{93.4\eb{0.2}} \\
 & GSM8K & \textbf{99.1\eb{0.1}} & \textbf{99.1\eb{0.1}} & \textbf{99.1\eb{0.2}} & \textbf{99.1\eb{0.1}} & 98.1\eb{0.3} & 98.1\eb{0.3} & \underline{98.8\eb{0.2}} & \underline{98.8\eb{0.2}} \\
 & OlympiadBench & 67.9\eb{0.4} & 67.9\eb{0.4} & \textbf{68.9\eb{0.1}} & \underline{68.6\eb{0.6}} & 66.3\eb{0.9} & 66.3\eb{0.9} & \underline{68.7\eb{0.1}} & \textbf{69.7\eb{0.8}} \\
 & Countdown & \underline{87.8\eb{1.0}} & \textbf{87.8\eb{1.0}} & \textbf{88.0\eb{1.3}} & \underline{87.6\eb{2.7}} & 81.6\eb{4.5} & 81.6\eb{4.5} & 86.4\eb{2.0} & 87.1\eb{1.9} \\
 & Knights \& Knaves & \textbf{100.0\eb{0.0}} & \textbf{100.0\eb{0.0}} & \textbf{100.0\eb{0.0}} & \underline{99.9\eb{0.2}} & \textbf{100.0\eb{0.0}} & \textbf{100.0\eb{0.0}} & \underline{99.9\eb{0.2}} & \textbf{100.0\eb{0.0}} \\
 & \emph{Mean} & \underline{89.6\eb{0.3}} & \underline{89.6\eb{0.3}} & \textbf{90.0\eb{0.3}} & 89.5\eb{0.5} & 87.7\eb{0.7} & 87.7\eb{0.7} & \underline{89.6\eb{0.5}} & \textbf{89.8\eb{0.6}} \\
\addlinespace[2pt]
Qwen2.5-3B & MATH500 & \underline{95.3\eb{0.7}} & \textbf{95.3\eb{0.7}} & 95.0\eb{0.4} & \underline{94.9\eb{0.1}} & 93.5\eb{0.8} & 93.5\eb{0.8} & \textbf{95.5\eb{0.5}} & \underline{94.9\eb{0.1}} \\
 & GSM8K & \textbf{99.3\eb{0.0}} & \textbf{99.3\eb{0.0}} & \textbf{99.3\eb{0.1}} & \textbf{99.3\eb{0.2}} & 98.5\eb{0.2} & \underline{98.5\eb{0.2}} & \underline{99.2\eb{0.2}} & \textbf{99.3\eb{0.2}} \\
 & OlympiadBench & \underline{75.1\eb{0.6}} & \underline{75.1\eb{0.6}} & \textbf{75.2\eb{1.2}} & 74.5\eb{0.7} & 72.2\eb{1.7} & 72.2\eb{1.7} & 74.5\eb{1.0} & \textbf{75.8\eb{1.2}} \\
 & Countdown & \textbf{89.1\eb{1.0}} & \underline{89.1\eb{1.0}} & \underline{88.4\eb{2.0}} & \textbf{89.3\eb{1.8}} & 80.9\eb{2.7} & 80.9\eb{2.7} & 87.3\eb{1.2} & 88.2\eb{1.0} \\
 & Knights \& Knaves & \underline{99.8\eb{0.2}} & \underline{99.8\eb{0.2}} & \textbf{100.0\eb{0.0}} & \textbf{100.0\eb{0.0}} & 99.7\eb{0.6} & 99.7\eb{0.6} & \textbf{100.0\eb{0.0}} & \textbf{100.0\eb{0.0}} \\
 & \emph{Mean} & \textbf{91.7\eb{0.1}} & \textbf{91.7\eb{0.1}} & \underline{91.6\eb{0.6}} & \underline{91.6\eb{0.5}} & 88.9\eb{1.0} & 88.9\eb{1.0} & 91.3\eb{0.4} & \underline{91.6\eb{0.1}} \\
\addlinespace[2pt]
Qwen2.5-7B & MATH500 & \textbf{96.3\eb{0.1}} & \textbf{96.3\eb{0.1}} & \underline{95.7\eb{0.2}} & \underline{95.8\eb{0.2}} & 95.5\eb{0.5} & 95.5\eb{0.5} & \textbf{96.3\eb{0.6}} & \textbf{96.3\eb{0.3}} \\
 & GSM8K & \underline{98.7\eb{0.0}} & \textbf{98.7\eb{0.0}} & \textbf{98.8\eb{0.2}} & \textbf{98.7\eb{0.1}} & 98.2\eb{0.1} & 98.2\eb{0.1} & 98.6\eb{0.1} & \underline{98.6\eb{0.1}} \\
 & OlympiadBench & \underline{78.8\eb{0.2}} & \underline{78.8\eb{0.2}} & \textbf{79.1\eb{0.3}} & \textbf{78.9\eb{0.6}} & 76.7\eb{0.7} & 76.7\eb{0.7} & 78.5\eb{0.5} & \textbf{78.9\eb{0.5}} \\
 & Countdown & \textbf{90.0\eb{1.8}} & \textbf{90.0\eb{1.8}} & \underline{89.3\eb{0.7}} & 88.2\eb{3.2} & 86.0\eb{2.0} & 86.0\eb{2.0} & 88.2\eb{0.8} & \underline{89.3\eb{0.7}} \\
 & Knights \& Knaves & \textbf{100.0\eb{0.0}} & \textbf{100.0\eb{0.0}} & \textbf{100.0\eb{0.0}} & \textbf{100.0\eb{0.0}} & \underline{99.9\eb{0.2}} & \underline{99.9\eb{0.2}} & \textbf{100.0\eb{0.0}} & \textbf{100.0\eb{0.0}} \\
 & \emph{Mean} & \textbf{92.8\eb{0.4}} & \textbf{92.8\eb{0.4}} & \underline{92.6\eb{0.2}} & 92.3\eb{0.8} & 91.2\eb{0.5} & 91.2\eb{0.5} & 92.3\eb{0.2} & \underline{92.6\eb{0.2}} \\
\addlinespace[2pt]
Llama-3.1-8B & MATH500 & 91.3\eb{0.5} & 91.3\eb{0.5} & \underline{92.1\eb{0.8}} & \underline{91.7\eb{0.7}} & 88.9\eb{0.3} & 88.9\eb{0.3} & \textbf{92.4\eb{0.3}} & \textbf{92.2\eb{0.6}} \\
 & GSM8K & \underline{99.2\eb{0.1}} & \textbf{99.2\eb{0.1}} & \underline{99.2\eb{0.1}} & \textbf{99.2\eb{0.1}} & 98.0\eb{0.5} & 98.0\eb{0.5} & \textbf{99.3\eb{0.2}} & \underline{99.1\eb{0.2}} \\
 & OlympiadBench & 63.6\eb{0.5} & 63.6\eb{0.5} & \underline{66.7\eb{1.3}} & \underline{64.8\eb{1.7}} & 63.5\eb{2.0} & 63.5\eb{2.0} & \textbf{68.6\eb{1.1}} & \textbf{68.0\eb{1.0}} \\
 & Countdown & 77.6\eb{1.5} & 77.6\eb{1.5} & 81.3\eb{2.4} & 78.4\eb{1.7} & \underline{81.6\eb{6.8}} & \underline{81.6\eb{6.8}} & \textbf{89.3\eb{1.8}} & \textbf{83.6\eb{1.4}} \\
 & Knights \& Knaves & \textbf{100.0\eb{0.0}} & \textbf{100.0\eb{0.0}} & \textbf{100.0\eb{0.0}} & \underline{99.9\eb{0.2}} & 99.4\eb{1.0} & 99.4\eb{1.0} & \underline{99.8\eb{0.4}} & 99.8\eb{0.4} \\
 & \emph{Mean} & 86.3\eb{0.3} & 86.3\eb{0.3} & \underline{87.9\eb{0.1}} & \underline{86.8\eb{0.6}} & 86.3\eb{2.0} & 86.3\eb{2.0} & \textbf{89.9\eb{0.7}} & \textbf{88.5\eb{0.2}} \\
\bottomrule
\end{tabular}

  \end{center}
\end{table}

\begin{table}[!htbp]
  \caption{Per-benchmark coverage (pass@pool \%) on the code suite, $K \in \{2,4\}$; mean\,$\pm$\,std over three seeds. Bold: best method per row and $K$; underlined: second best.}
  \label{tab:cov_code_perbench}
  \begin{center}
  \scriptsize
  \setlength{\tabcolsep}{2.5pt}
  \begin{tabular}{llcccccccc}
\toprule
 & & \multicolumn{2}{c}{Base} & \multicolumn{2}{c}{\singlem{}} & \multicolumn{2}{c}{\single{}} & \multicolumn{2}{c}{Thicket} \\
\cmidrule(lr){3-4} \cmidrule(lr){5-6} \cmidrule(lr){7-8} \cmidrule(lr){9-10}
Model & Benchmark ($K{=}$) & 2 & 4 & 2 & 4 & 2 & 4 & 2 & 4 \\
\midrule
Qwen2.5-1.5B & MBPP & \textbf{93.4\eb{0.0}} & \textbf{93.4\eb{0.0}} & 90.6\eb{2.4} & 92.4\eb{1.0} & 88.1\eb{1.6} & 88.1\eb{1.6} & \underline{91.4\eb{0.9}} & \underline{92.7\eb{0.3}} \\
 & HumanEval & \textbf{96.7\eb{0.9}} & \textbf{96.7\eb{0.9}} & 92.5\eb{3.0} & \underline{93.9\eb{0.6}} & 88.8\eb{1.4} & 88.8\eb{1.4} & \underline{92.9\eb{1.4}} & \underline{93.9\eb{0.6}} \\
 & LeetCode & \textbf{26.5\eb{0.3}} & \textbf{26.5\eb{0.3}} & \underline{24.0\eb{3.7}} & 24.4\eb{2.1} & 20.5\eb{2.6} & 20.5\eb{2.6} & 23.4\eb{2.8} & \underline{25.1\eb{1.1}} \\
 & \emph{Mean} & \textbf{72.2\eb{0.4}} & \textbf{72.2\eb{0.4}} & 69.0\eb{1.8} & 70.2\eb{0.9} & 65.8\eb{1.4} & 65.8\eb{1.4} & \underline{69.2\eb{1.4}} & \underline{70.6\eb{0.6}} \\
\addlinespace[2pt]
Qwen2.5-3B & MBPP & \textbf{95.2\eb{0.5}} & \textbf{95.2\eb{0.5}} & \underline{93.0\eb{2.8}} & 92.9\eb{1.8} & 90.8\eb{1.1} & 90.8\eb{1.1} & 92.6\eb{1.3} & \underline{93.1\eb{1.8}} \\
 & HumanEval & \textbf{97.6\eb{0.6}} & \textbf{97.6\eb{0.6}} & 96.5\eb{1.5} & 96.1\eb{1.5} & 95.3\eb{0.9} & 95.3\eb{0.9} & \underline{97.4\eb{0.9}} & \underline{96.7\eb{0.7}} \\
 & LeetCode & \textbf{34.1\eb{2.4}} & \textbf{34.1\eb{2.4}} & \underline{31.7\eb{2.1}} & 30.4\eb{2.1} & 26.0\eb{1.5} & 26.0\eb{1.5} & 30.4\eb{2.1} & \underline{32.2\eb{1.8}} \\
 & \emph{Mean} & \textbf{75.6\eb{1.1}} & \textbf{75.6\eb{1.1}} & \underline{73.7\eb{2.0}} & 73.1\eb{1.7} & 70.7\eb{0.5} & 70.7\eb{0.5} & 73.5\eb{1.4} & \underline{74.0\eb{1.4}} \\
\addlinespace[2pt]
Qwen2.5-7B & MBPP & \textbf{93.1\eb{0.1}} & \textbf{93.1\eb{0.1}} & 92.0\eb{0.9} & 92.1\eb{1.2} & 90.1\eb{0.9} & 90.1\eb{0.9} & \underline{93.0\eb{0.5}} & \underline{92.6\eb{0.5}} \\
 & HumanEval & \textbf{96.7\eb{0.4}} & \textbf{96.7\eb{0.4}} & 94.7\eb{0.9} & \underline{96.5\eb{0.4}} & 94.9\eb{0.4} & 94.9\eb{0.4} & \underline{96.3\eb{0.6}} & 96.1\eb{0.9} \\
 & LeetCode & \textbf{39.0\eb{0.9}} & \textbf{39.0\eb{0.9}} & 36.3\eb{0.7} & 38.3\eb{1.1} & 34.2\eb{0.8} & 34.2\eb{0.8} & \underline{37.9\eb{0.3}} & \underline{38.9\eb{0.9}} \\
 & \emph{Mean} & \textbf{76.3\eb{0.2}} & \textbf{76.3\eb{0.2}} & 74.3\eb{0.5} & 75.7\eb{0.7} & 73.1\eb{0.4} & 73.1\eb{0.4} & \underline{75.7\eb{0.2}} & \underline{75.9\eb{0.2}} \\
\addlinespace[2pt]
Llama-3.1-8B & MBPP & \textbf{93.4\eb{0.5}} & \textbf{93.4\eb{0.5}} & 89.2\eb{0.5} & \underline{90.9\eb{1.0}} & 88.4\eb{1.6} & 88.4\eb{1.6} & \underline{90.7\eb{0.8}} & 90.6\eb{0.8} \\
 & HumanEval & \textbf{93.1\eb{0.7}} & \textbf{93.1\eb{0.7}} & \underline{90.7\eb{2.7}} & 91.7\eb{1.5} & 90.4\eb{0.9} & 90.4\eb{0.9} & 90.0\eb{2.0} & \underline{91.9\eb{0.4}} \\
 & LeetCode & \textbf{39.0\eb{0.4}} & \underline{39.0\eb{0.4}} & \underline{38.0\eb{2.1}} & 38.7\eb{1.5} & 36.7\eb{3.2} & 36.7\eb{3.2} & \textbf{39.0\eb{1.9}} & \textbf{40.6\eb{0.5}} \\
 & \emph{Mean} & \textbf{75.2\eb{0.2}} & \textbf{75.2\eb{0.2}} & 72.6\eb{1.7} & 73.8\eb{1.1} & 71.8\eb{1.9} & 71.8\eb{1.9} & \underline{73.3\eb{1.4}} & \underline{74.4\eb{0.5}} \\
\bottomrule
\end{tabular}

  \end{center}
\end{table}

\begin{table}[!htbp]
  \caption{Per-benchmark coverage (pass@pool \%) on the code suite, $K \in \{8,16\}$ (format of Table~\ref{tab:cov_code_perbench}).}
  \label{tab:cov_code_perbench_b}
  \begin{center}
  \scriptsize
  \setlength{\tabcolsep}{2.5pt}
  \begin{tabular}{llcccccccc}
\toprule
 & & \multicolumn{2}{c}{Base} & \multicolumn{2}{c}{\singlem{}} & \multicolumn{2}{c}{\single{}} & \multicolumn{2}{c}{Thicket} \\
\cmidrule(lr){3-4} \cmidrule(lr){5-6} \cmidrule(lr){7-8} \cmidrule(lr){9-10}
Model & Benchmark ($K{=}$) & 8 & 16 & 8 & 16 & 8 & 16 & 8 & 16 \\
\midrule
Qwen2.5-1.5B & MBPP & \underline{93.4\eb{0.0}} & \underline{93.4\eb{0.0}} & \textbf{93.8\eb{0.1}} & \textbf{93.7\eb{0.8}} & 88.1\eb{1.6} & 88.1\eb{1.6} & 93.3\eb{0.1} & 93.0\eb{0.5} \\
 & HumanEval & \textbf{96.7\eb{0.9}} & \textbf{96.7\eb{0.9}} & \textbf{96.7\eb{0.7}} & \underline{96.1\eb{1.5}} & 88.8\eb{1.4} & 88.8\eb{1.4} & \underline{94.1\eb{1.4}} & 93.5\eb{0.9} \\
 & LeetCode & \textbf{26.5\eb{0.3}} & \textbf{26.5\eb{0.3}} & \underline{25.6\eb{0.9}} & \underline{25.9\eb{0.4}} & 20.5\eb{2.6} & 20.5\eb{2.6} & 24.3\eb{0.3} & 24.6\eb{1.5} \\
 & \emph{Mean} & \textbf{72.2\eb{0.4}} & \textbf{72.2\eb{0.4}} & \underline{72.0\eb{0.5}} & \underline{71.9\eb{0.3}} & 65.8\eb{1.4} & 65.8\eb{1.4} & 70.6\eb{0.5} & 70.3\eb{0.3} \\
\addlinespace[2pt]
Qwen2.5-3B & MBPP & \textbf{95.2\eb{0.5}} & 95.2\eb{0.5} & \underline{94.8\eb{0.5}} & \textbf{95.6\eb{0.2}} & 90.8\eb{1.1} & 90.8\eb{1.1} & 94.2\eb{0.9} & \underline{95.4\eb{0.7}} \\
 & HumanEval & 97.6\eb{0.6} & 97.6\eb{0.6} & \underline{97.8\eb{0.7}} & \textbf{98.2\eb{0.6}} & 95.3\eb{0.9} & 95.3\eb{0.9} & \textbf{98.0\eb{0.4}} & \underline{97.8\eb{0.4}} \\
 & LeetCode & \textbf{34.1\eb{2.4}} & 34.1\eb{2.4} & \underline{33.8\eb{1.8}} & \textbf{35.4\eb{2.0}} & 26.0\eb{1.5} & 26.0\eb{1.5} & 33.5\eb{0.7} & \underline{34.6\eb{0.4}} \\
 & \emph{Mean} & \textbf{75.6\eb{1.1}} & 75.6\eb{1.1} & \underline{75.4\eb{0.8}} & \textbf{76.4\eb{0.9}} & 70.7\eb{0.5} & 70.7\eb{0.5} & 75.2\eb{0.4} & \underline{75.9\eb{0.3}} \\
\addlinespace[2pt]
Qwen2.5-7B & MBPP & \underline{93.1\eb{0.1}} & \underline{93.1\eb{0.1}} & 92.9\eb{0.4} & \textbf{93.2\eb{0.5}} & 90.1\eb{0.9} & 90.1\eb{0.9} & \textbf{93.4\eb{0.4}} & 92.7\eb{0.1} \\
 & HumanEval & \textbf{96.7\eb{0.4}} & \underline{96.7\eb{0.4}} & 96.1\eb{0.4} & \underline{96.7\eb{0.4}} & 94.9\eb{0.4} & 94.9\eb{0.4} & \underline{96.5\eb{0.9}} & \textbf{97.0\eb{0.0}} \\
 & LeetCode & \underline{39.0\eb{0.9}} & 39.0\eb{0.9} & \textbf{39.8\eb{1.8}} & \underline{39.3\eb{1.0}} & 34.2\eb{0.8} & 34.2\eb{0.8} & 38.9\eb{0.5} & \textbf{39.5\eb{1.2}} \\
 & \emph{Mean} & \textbf{76.3\eb{0.2}} & \underline{76.3\eb{0.2}} & \textbf{76.3\eb{0.4}} & \textbf{76.4\eb{0.2}} & \underline{73.1\eb{0.4}} & 73.1\eb{0.4} & \textbf{76.3\eb{0.4}} & \textbf{76.4\eb{0.4}} \\
\addlinespace[2pt]
Llama-3.1-8B & MBPP & \textbf{93.4\eb{0.5}} & \underline{93.4\eb{0.5}} & \underline{92.8\eb{0.5}} & \textbf{93.6\eb{0.3}} & 88.4\eb{1.6} & 88.4\eb{1.6} & 92.3\eb{0.2} & 92.1\eb{0.7} \\
 & HumanEval & \textbf{93.1\eb{0.7}} & \underline{93.1\eb{0.7}} & \underline{92.9\eb{0.4}} & \textbf{93.3\eb{0.0}} & 90.4\eb{0.9} & 90.4\eb{0.9} & 92.3\eb{0.9} & 92.3\eb{0.4} \\
 & LeetCode & \underline{39.0\eb{0.4}} & 39.0\eb{0.4} & \textbf{39.2\eb{1.0}} & \underline{39.3\eb{0.7}} & 36.7\eb{3.2} & 36.7\eb{3.2} & 38.5\eb{1.5} & \textbf{40.1\eb{0.7}} \\
 & \emph{Mean} & \textbf{75.2\eb{0.2}} & \underline{75.2\eb{0.2}} & \underline{75.0\eb{0.4}} & \textbf{75.4\eb{0.1}} & 71.8\eb{1.9} & 71.8\eb{1.9} & 74.3\eb{0.1} & 74.8\eb{0.5} \\
\bottomrule
\end{tabular}

  \end{center}
\end{table}

\clearpage

\section{Voter accuracy, error correlation, and a margin decomposition}
\label{app:corr}

\begin{figure}[!htbp]
    \centering
    \includegraphics[width=0.95\linewidth]{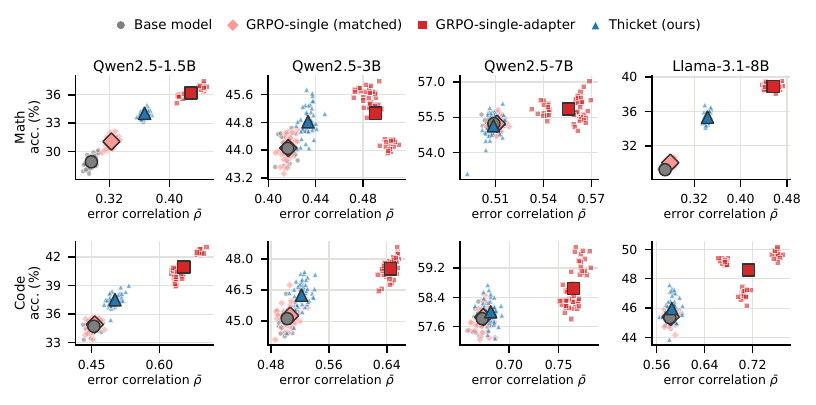}
    \caption{\textbf{Voter accuracy vs.\ error correlation at $K{=}16$} (per-voter macro means over each suite's benchmarks, voters of all three seeds pooled; large markers are method means). The fully trained \single{} moves up and far to the right of the base model. The step-matched control stays close to the base model in both correlation and accuracy. Thicket members add less correlation than \single{} in every panel, and gain accuracy over the base model in all but Qwen2.5-7B math ($-0.1$).}
    \label{fig:mechanism}
\end{figure}

\begin{figure}[!htbp]
    \centering
    \includegraphics[width=0.95\linewidth]{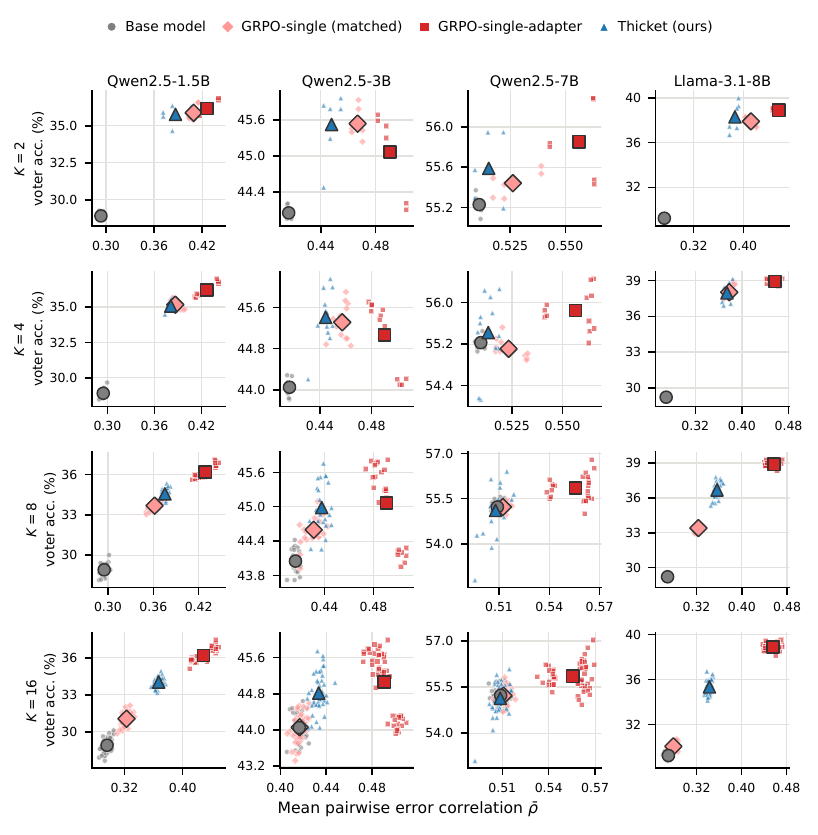}
    \caption{Complete plot set, math suite: per-voter accuracy vs.\ $\corr$ for $K \in \{2,4,8,16\}$ (rows) and all models (columns); voters of all three seeds pooled; large markers are method means.}
    \label{fig:corr_grid_math}
\end{figure}

\begin{figure}[!htbp]
    \centering
    \includegraphics[width=0.95\linewidth]{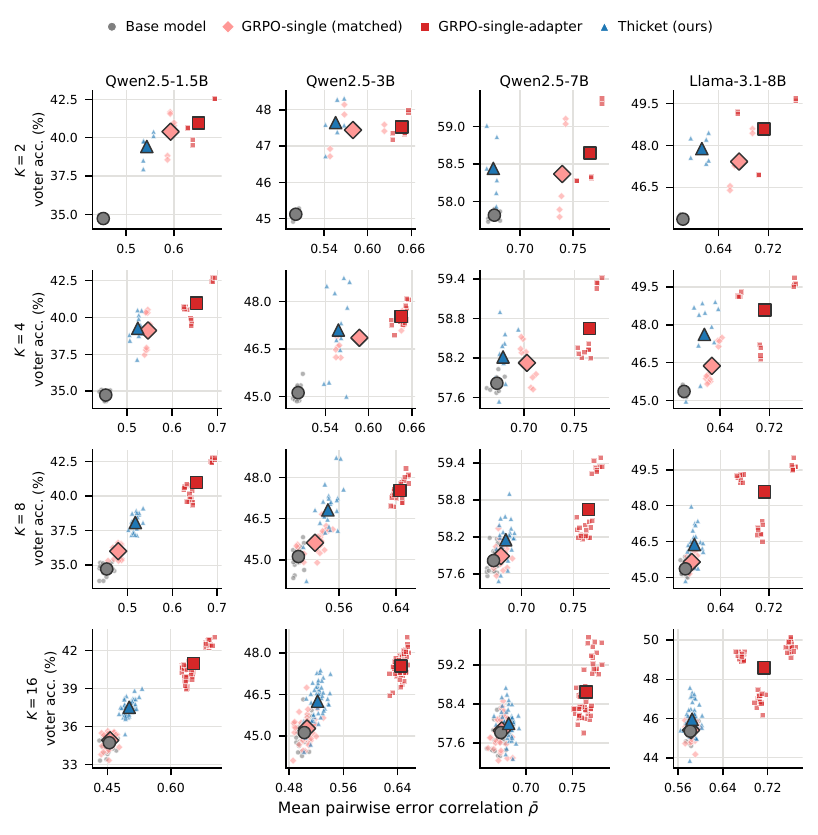}
    \caption{Complete plot set, code suite (protocol as in Figure~\ref{fig:corr_grid_math}).}
    \label{fig:corr_grid_code}
\end{figure}

Figure~\ref{fig:mechanism} places every voter at $K{=}16$ in the accuracy--correlation plane, and Table~\ref{tab:corr16} summarizes it; Figures~\ref{fig:corr_grid_math} and \ref{fig:corr_grid_code} give the complete plot set for $K \in \{2,4,8,16\}$ on both suites, with the voters of all three seeds pooled in each panel. Per-benchmark $\corr$ is computed over each voter's $\nicefrac{160}{K}$ stored completions; suite-level values are per-voter macro means over the suite's benchmarks.

\textbf{Where does the thicket's margin come from?} Table~\ref{tab:decomp} splits the thicket's $K{=}16$ margin over \single{} at $160$ votes into the vote gain it keeps over the base and the damage it avoids. On Qwen2.5-3B and Llama, most of the margin is avoided damage ($+4.8$ and $+4.1$ p.p.). On Qwen2.5-7B the avoided damage ($+1.7$) exceeds the whole margin ($+1.2$), because the thicket itself trails the base by $0.5$. On Qwen2.5-1.5B, where the fully trained adapter's vote merely matches the base, the margin ($+2.6$) comes entirely from retained gain ($+2.9$, offset by $-0.3$). The retained gain in turn tracks how much RLVR can add at all. Where full training lifts pass@1 by $+7.3$ and $+9.7$ p.p.\ (Qwen2.5-1.5B and Llama), the thicket beats the base ($+2.9$ and $+1.4$). Where it lifts pass@1 by only $+1.0$ and $+0.7$ (Qwen2.5-3B and 7B), the thicket roughly matches the base ($+0.3$ and $-0.5$). On these four models, a thicket converts RLVR's accuracy gain into a vote gain when there is one, and otherwise shields the base model's vote from most of the damage that concentrated training inflicts.

\begin{table}[!htbp]
  \caption{\textbf{Decomposing the thicket's vote margin over \single{}} at $K{=}16$ and $160$ votes (math macro, mean over three seeds). $G = (\text{thicket} - \text{base}) + (\text{base} - \text{\single{}})$: the RLVR gain the thicket retains, plus the damage that concentrated training inflicts and the thicket avoids. $H$ and $\Delta\corr$ are the pass@1 and error-correlation changes that full single-adapter training causes relative to the base.}
  \label{tab:decomp}
  \begin{center}
  \small
  \setlength{\tabcolsep}{5pt}
  \begin{tabular}{lccccc}
\toprule
 & \multicolumn{2}{c}{What full RLVR does} & \multicolumn{3}{c}{Thicket margin (MV, p.p.)} \\
\cmidrule(lr){2-3} \cmidrule(lr){4-6}
Model & $H$ (pass@1) & $\Delta\corr$ & thicket$-$base & base$-$\single{} & $G$ \\
\midrule
Qwen2.5-1.5B & +7.3 & +0.13 & +2.9 & $-$0.3 & +2.6 \\
Qwen2.5-3B & +1.0 & +0.07 & +0.3 & +4.8 & +5.1 \\
Qwen2.5-7B & +0.7 & +0.05 & $-$0.5 & +1.7 & +1.2 \\
Llama-3.1-8B & +9.7 & +0.19 & +1.4 & +4.1 & +5.5 \\
\bottomrule
\end{tabular}

  \end{center}
\end{table}

\begin{table}[!htbp]
  \caption{Mean voter accuracy (single completion, \%) and mean pairwise error correlation $\corr$ at $K{=}16$; macro means over each suite's benchmarks, mean\,$\pm$\,std over three seeds. Matched: \singlem{}; Single: \single{}.}
  \label{tab:corr16}
  \begin{center}
  \scriptsize
  \setlength{\tabcolsep}{3pt}
  \begin{tabular}{llcccccccc}
\toprule
 & & \multicolumn{4}{c}{Voter accuracy (\%)} & \multicolumn{4}{c}{Error correlation $\corr$} \\
\cmidrule(lr){3-6} \cmidrule(lr){7-10}
Suite & Model & Base & Matched & Single & Thicket & Base & Matched & Single & Thicket \\
\midrule
Math & Qwen2.5-1.5B & 28.9\eb{0.1} & 31.1\eb{0.1} & 36.2\eb{0.5} & 34.0\eb{0.1} & 0.30\eb{0.00} & 0.32\eb{0.00} & 0.43\eb{0.01} & 0.37\eb{0.00} \\
 & Qwen2.5-3B & 44.1\eb{0.1} & 44.1\eb{0.2} & 45.1\eb{0.8} & 44.8\eb{0.1} & 0.42\eb{0.00} & 0.42\eb{0.00} & 0.49\eb{0.01} & 0.43\eb{0.00} \\
 & Qwen2.5-7B & 55.2\eb{0.1} & 55.2\eb{0.0} & 55.9\eb{0.4} & 55.1\eb{0.0} & 0.51\eb{0.00} & 0.51\eb{0.00} & 0.56\eb{0.01} & 0.51\eb{0.00} \\
 & Llama-3.1-8B & 29.2\eb{0.0} & 30.0\eb{0.2} & 38.9\eb{0.2} & 35.3\eb{0.0} & 0.27\eb{0.00} & 0.28\eb{0.00} & 0.46\eb{0.01} & 0.34\eb{0.00} \\
\midrule
Code & Qwen2.5-1.5B & 34.7\eb{0.1} & 34.9\eb{0.2} & 41.0\eb{1.5} & 37.5\eb{0.2} & 0.45\eb{0.00} & 0.46\eb{0.00} & 0.65\eb{0.03} & 0.50\eb{0.01} \\
 & Qwen2.5-3B & 45.1\eb{0.1} & 45.3\eb{0.1} & 47.5\eb{0.4} & 46.3\eb{0.1} & 0.50\eb{0.00} & 0.51\eb{0.01} & 0.65\eb{0.01} & 0.52\eb{0.00} \\
 & Qwen2.5-7B & 57.8\eb{0.1} & 57.9\eb{0.1} & 58.6\eb{0.6} & 58.0\eb{0.1} & 0.67\eb{0.00} & 0.67\eb{0.00} & 0.77\eb{0.01} & 0.68\eb{0.00} \\
 & Llama-3.1-8B & 45.4\eb{0.1} & 45.4\eb{0.2} & 48.6\eb{1.4} & 46.0\eb{0.2} & 0.58\eb{0.00} & 0.58\eb{0.00} & 0.71\eb{0.04} & 0.59\eb{0.01} \\
\bottomrule
\end{tabular}

  \end{center}
\end{table}

\clearpage

\section{The vote budget: from 16 to 160 votes}
\label{app:budget}

\begin{figure}[!htbp]
    \centering
    \includegraphics[width=0.95\linewidth]{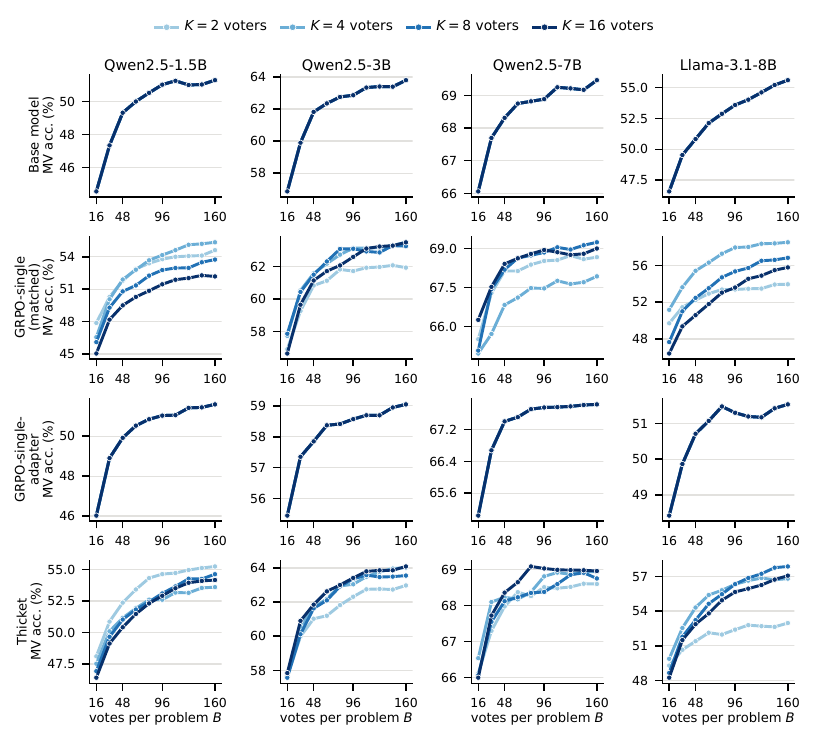}
    \caption{Majority-vote accuracy (math macro, mean over three seeds) as each method votes over its first $B$ completions, $B = 16, 32, \dots, 160$, per voter count $K$ (line shade) and method (rows). The base-model and \single{} curves are identical at every $K$, so only the $K{=}16$ curve is visible in their rows. No bands are drawn here. Figure~\ref{fig:budget_methods} draws $\pm 1$ std bands.}
    \label{fig:budget_curves}
\end{figure}

\begin{figure}[!htbp]
    \centering
    \includegraphics[width=0.95\linewidth]{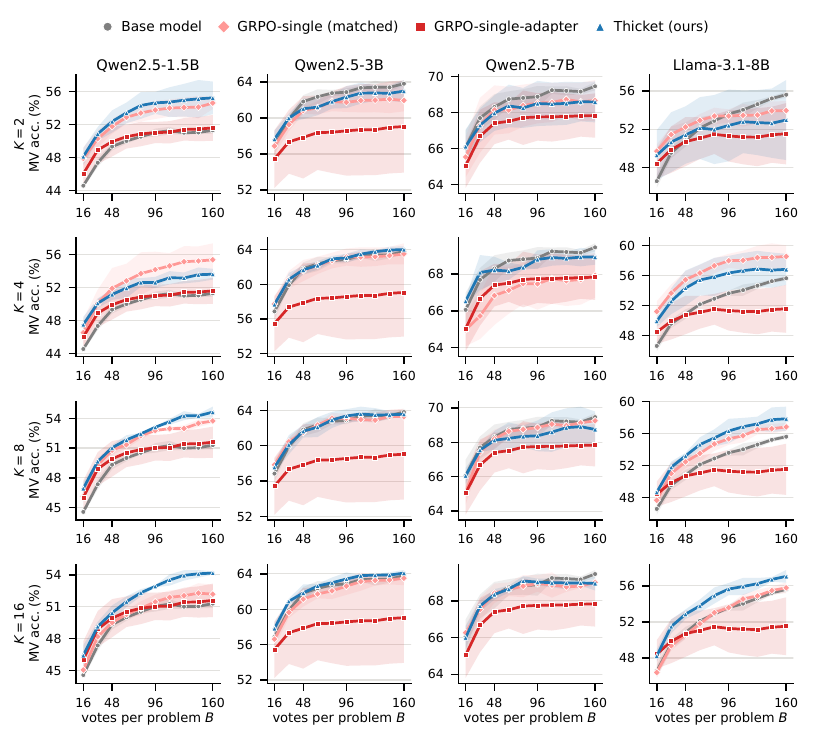}
    \caption{The same curves as Figure~\ref{fig:budget_curves}, with the four methods overlaid in each panel, per voter count $K$ (rows) and model (columns). Bands: $\pm 1$ std over seeds.}
    \label{fig:budget_methods}
\end{figure}

Section~\ref{sec:budget} summarizes these results by mean margins. Here every method votes over its first $B$ completions, $B \in \{16, 32, \dots, 160\}$, under the first-$B$ convention of Appendix~\ref{app:eval}. Table~\ref{tab:budget} gives the mean margins of Figure~\ref{fig:budget} at seven budgets, together with each method's mean vote. Figure~\ref{fig:budget_curves} shows each method's curves by voter count $K$, and Figure~\ref{fig:budget_methods} overlays the four methods in each (model, $K$) panel.

Three patterns stand out. First, the base-model and \single{} curves are identical at every $K$, because their first $B$ completions do not depend on $K$. Second, the fully trained \single{} saturates fastest. Its vote rises by $3.78$ p.p.\ from $16$ to $160$ votes, against $6.50$ for the base model (means of the per-setting gains). Along the single adapter's run the gain shrinks as training continues past step $50$: it is $+6.55$ and $+6.58$ p.p.\ at steps $25$ and $50$ (\singlem{} at $K{=}16$ and $8$), $+6.18$ at step $125$, $+4.78$ at step $250$, and $+3.78$ at the final checkpoint. Third, larger thickets gain more from the same extra votes: $+4.68$ p.p.\ at $K{=}2$, $+5.43$ at $K{=}4$, $+6.35$ at $K{=}8$, and $+6.48$ at $K{=}16$ (means over models of the per-model gains).

\begin{table}[!htbp]
  \caption{\textbf{The vote budget} (math majority vote over each method's first $B$ completions; three-seed means). Upper rows: mean margin (p.p.) over the $16$ (model, $K$) settings, each the difference of displayed means, with the number of settings where it is positive in parentheses. Lower rows: each method's majority vote, averaged over the $16$ settings, $\pm$ the std over seeds of that average. $B{=}160$ is Table~\ref{tab:mv}, and $B{=}16$ is Table~\ref{tab:mv16}. Figure~\ref{fig:budget} plots all ten budgets. Base and \single{} do not depend on $K$, so the count for \single{} $-$ Base moves in steps of four (one model at a time).}
  \label{tab:budget}
  \begin{center}
  \footnotesize
  \setlength{\tabcolsep}{3.5pt}
  \resizebox{\linewidth}{!}{\begin{tabular}{lccccccc}
\toprule
 & \multicolumn{7}{c}{Votes per problem $B$ (first $B$ responses of each method)} \\
\cmidrule(lr){2-8}
 & 16 & 32 & 48 & 64 & 96 & 128 & 160 \\
\midrule
\multicolumn{8}{l}{\emph{Mean margin (p.p.) over the 16 (model, $K$) settings (settings with a positive margin)}} \\
\single{} $-$ Base & +0.15\,(8) & $-$0.40\,(8) & $-$1.08\,(4) & $-$1.43\,(4) & $-$1.93\,(0) & $-$2.28\,(4) & $-$2.58\,(4) \\
Thicket $-$ \single{} & +1.33\,(15) & +1.68\,(16) & +2.01\,(16) & +2.25\,(16) & +2.86\,(16) & +3.26\,(16) & +3.28\,(16) \\
Thicket $-$ Base & +1.48\,(13) & +1.28\,(13) & +0.93\,(10) & +0.83\,(8) & +0.94\,(11) & +0.98\,(10) & +0.71\,(9) \\
Thicket $-$ \singlem{} & +0.56\,(11) & +0.47\,(11) & +0.28\,(10) & +0.28\,(9) & +0.23\,(9) & +0.34\,(11) & +0.27\,(10) \\
\midrule
\multicolumn{8}{l}{\emph{Majority vote (\%), mean over the 16 settings}} \\
Base & 53.6\eb{0.4} & 56.1\eb{0.3} & 57.6\eb{0.4} & 58.3\eb{0.3} & 59.1\eb{0.2} & 59.6\eb{0.0} & 60.1\eb{0.2} \\
\singlem{} & 54.5\eb{0.5} & 56.9\eb{0.5} & 58.2\eb{0.6} & 58.8\eb{0.4} & 59.8\eb{0.4} & 60.2\eb{0.4} & 60.5\eb{0.4} \\
\single{} & 53.7\eb{1.3} & 55.7\eb{1.5} & 56.5\eb{1.7} & 56.9\eb{1.6} & 57.2\eb{1.8} & 57.3\eb{2.0} & 57.5\eb{2.2} \\
Thicket & 55.0\eb{0.4} & 57.4\eb{0.2} & 58.5\eb{0.1} & 59.1\eb{0.2} & 60.0\eb{0.2} & 60.5\eb{0.2} & 60.8\eb{0.3} \\
\bottomrule
\end{tabular}
}
  \end{center}
\end{table}

\textbf{A 16-vote replication.} Tables~\ref{tab:mv16} and~\ref{tab:cov16} repeat Tables~\ref{tab:mv} and~\ref{tab:cov} at $B{=}16$, the smallest budget that splits evenly across every $K \in \{2,4,8,16\}$. A thicket member contributes its first $\nicefrac{16}{K}$ completions ($8/4/2/1$ for $K{=}2/4/8/16$), and every single policy the first $16$ completions of its pool, those drawn with its first $16$ sampling seeds. At $16$ votes the thicket beats \single{} in $15$ of $16$ settings (mean margin $+1.33$ p.p.), the base in $13$ (tying $2$; $+1.48$), and \singlem{} in $11$ (tying $1$; $+0.56$). The fully trained \single{} votes above the base on Qwen2.5-1.5B ($+1.4$) and Llama ($+1.8$), and below it on Qwen2.5-3B ($-1.5$) and Qwen2.5-7B ($-1.1$). For \single{}, coverage tells the same story as at $160$ votes. Thickets cover more than \single{} in all $32$ (model, suite, $K$) settings (mean margins $+2.02$ p.p.\ on math and $+3.03$ on code). \single{} covers less than the base in seven of eight model--suite settings, and more only on Llama math ($+3.2$), where it ties the base at $160$ votes. Thickets cover more than the base in $9$ of $16$ settings on each suite. That is more often than at $160$ votes, where they are below the base in $23$ of $32$, in line with their shrinking lead over the base.

\begin{table}[!htbp]
  \caption{\textbf{Majority-vote accuracy on the math suite at $16$ votes} (canonical-class votes; macro means (\%) over five benchmarks; mean\,$\pm$\,std over three seeds). A thicket member contributes its first $\nicefrac{16}{K}$ completions. Every other method contributes the first $16$ completions of its pool, so the base and \single{} rows are constant across $K$ by construction. Bold: best method per model and $K$ by displayed mean (ties bold both); underlined: second best.}
  \label{tab:mv16}
  \begin{center}
  \small
  \setlength{\tabcolsep}{5pt}
  \begin{tabular}{llcccc}
\toprule
Model & Method & $K{=}2$ & $K{=}4$ & $K{=}8$ & $K{=}16$ \\
\midrule
Qwen2.5-1.5B & Base & 44.6\eb{0.3} & 44.6\eb{0.3} & 44.6\eb{0.3} & 44.6\eb{0.3} \\
 & \singlem{} & \underline{47.9\eb{1.4}} & \underline{46.6\eb{1.7}} & \underline{46.1\eb{0.4}} & 45.0\eb{0.1} \\
 & \single{} & 46.0\eb{1.7} & 46.0\eb{1.7} & 46.0\eb{1.7} & \underline{46.0\eb{1.7}} \\
 & Thicket & \textbf{48.1\eb{1.0}} & \textbf{47.5\eb{0.9}} & \textbf{46.9\eb{0.6}} & \textbf{46.4\eb{1.0}} \\
\addlinespace[2pt]
Qwen2.5-3B & Base & \underline{56.9\eb{0.9}} & \underline{56.9\eb{0.9}} & 56.9\eb{0.9} & \underline{56.9\eb{0.9}} \\
 & \singlem{} & \underline{56.9\eb{1.0}} & \textbf{57.7\eb{0.8}} & \textbf{57.8\eb{0.7}} & 56.6\eb{0.9} \\
 & \single{} & 55.4\eb{3.2} & 55.4\eb{3.2} & 55.4\eb{3.2} & 55.4\eb{3.2} \\
 & Thicket & \textbf{57.6\eb{1.0}} & \textbf{57.7\eb{0.2}} & \underline{57.6\eb{0.6}} & \textbf{57.8\eb{0.6}} \\
\addlinespace[2pt]
Qwen2.5-7B & Base & \textbf{66.1\eb{0.9}} & \underline{66.1\eb{0.9}} & \textbf{66.1\eb{0.9}} & \underline{66.1\eb{0.9}} \\
 & \singlem{} & \underline{65.5\eb{1.1}} & 65.0\eb{0.9} & \underline{65.1\eb{0.7}} & \textbf{66.3\eb{0.4}} \\
 & \single{} & 65.0\eb{1.2} & 65.0\eb{1.2} & 65.0\eb{1.2} & 65.0\eb{1.2} \\
 & Thicket & \textbf{66.1\eb{0.9}} & \textbf{66.5\eb{0.2}} & \textbf{66.1\eb{0.5}} & 66.0\eb{0.6} \\
\addlinespace[2pt]
Llama-3.1-8B & Base & 46.6\eb{0.7} & 46.6\eb{0.7} & 46.6\eb{0.7} & 46.6\eb{0.7} \\
 & \singlem{} & \textbf{49.7\eb{0.7}} & \textbf{51.2\eb{1.4}} & 47.7\eb{0.8} & 46.4\eb{0.6} \\
 & \single{} & 48.4\eb{1.6} & 48.4\eb{1.6} & \underline{48.4\eb{1.6}} & \textbf{48.4\eb{1.6}} \\
 & Thicket & \underline{49.3\eb{2.6}} & \underline{49.9\eb{0.8}} & \textbf{48.7\eb{0.2}} & \underline{48.2\eb{0.7}} \\
\bottomrule
\end{tabular}

  \end{center}
\end{table}

\begin{table}[!htbp]
  \caption{\textbf{Coverage at $16$ votes} (at least one correct completion among the $16$ pooled, with the first-$16$ convention of Table~\ref{tab:mv16}) on the math and code suites; macro means (\%); mean\,$\pm$\,std over three seeds. Bold: best method per model, suite, and $K$; underlined: second best.}
  \label{tab:cov16}
  \begin{center}
  \scriptsize
  \setlength{\tabcolsep}{2.6pt}
  \begin{tabular}{llcccccccc}
\toprule
 & & \multicolumn{4}{c}{Coverage, math} & \multicolumn{4}{c}{Coverage, code} \\
\cmidrule(lr){3-6} \cmidrule(lr){7-10}
Model & Method ($K{=}$) & 2 & 4 & 8 & 16 & 2 & 4 & 8 & 16 \\
\midrule
Qwen2.5-1.5B & Base & 73.1\eb{0.4} & 73.1\eb{0.4} & 73.1\eb{0.4} & 73.1\eb{0.4} & \underline{61.3\eb{1.1}} & 61.3\eb{1.1} & 61.3\eb{1.1} & 61.3\eb{1.1} \\
 & \singlem{} & \textbf{74.7\eb{0.8}} & \textbf{74.4\eb{1.5}} & \underline{73.7\eb{0.7}} & \underline{73.3\eb{0.5}} & \textbf{61.4\eb{2.2}} & \underline{61.4\eb{1.1}} & \textbf{62.5\eb{0.8}} & \underline{61.9\eb{0.9}} \\
 & \single{} & 72.8\eb{1.5} & 72.8\eb{1.5} & 72.8\eb{1.5} & 72.8\eb{1.5} & 58.2\eb{0.8} & 58.2\eb{0.8} & 58.2\eb{0.8} & 58.2\eb{0.8} \\
 & Thicket & \underline{74.5\eb{0.6}} & \underline{74.2\eb{0.8}} & \textbf{74.3\eb{0.6}} & \textbf{73.9\eb{0.7}} & \textbf{61.4\eb{1.9}} & \textbf{62.1\eb{1.1}} & \underline{62.4\eb{0.7}} & \textbf{63.5\eb{0.3}} \\
\addlinespace[2pt]
Qwen2.5-3B & Base & \textbf{80.6\eb{0.0}} & \textbf{80.6\eb{0.0}} & \textbf{80.6\eb{0.0}} & \underline{80.6\eb{0.0}} & \textbf{68.1\eb{0.5}} & \textbf{68.1\eb{0.5}} & \underline{68.1\eb{0.5}} & \underline{68.1\eb{0.5}} \\
 & \singlem{} & 78.8\eb{0.8} & 79.8\eb{0.9} & \underline{80.5\eb{0.3}} & \textbf{81.0\eb{0.2}} & \underline{67.3\eb{2.1}} & 65.5\eb{1.5} & 67.9\eb{0.4} & 68.0\eb{1.2} \\
 & \single{} & 76.5\eb{1.9} & 76.5\eb{1.9} & 76.5\eb{1.9} & 76.5\eb{1.9} & 64.9\eb{0.9} & 64.9\eb{0.9} & 64.9\eb{0.9} & 64.9\eb{0.9} \\
 & Thicket & \underline{78.9\eb{0.9}} & \underline{80.1\eb{0.8}} & 80.1\eb{0.6} & 80.4\eb{0.2} & 66.8\eb{0.6} & \underline{67.8\eb{0.3}} & \textbf{68.8\eb{1.0}} & \textbf{68.6\eb{1.0}} \\
\addlinespace[2pt]
Qwen2.5-7B & Base & \textbf{84.6\eb{0.3}} & \textbf{84.6\eb{0.3}} & \underline{84.6\eb{0.3}} & \underline{84.6\eb{0.3}} & \textbf{71.6\eb{0.3}} & \textbf{71.6\eb{0.3}} & \textbf{71.6\eb{0.3}} & 71.6\eb{0.3} \\
 & \singlem{} & 83.8\eb{0.7} & 83.6\eb{0.3} & 84.3\eb{0.4} & \textbf{84.8\eb{0.2}} & 69.8\eb{0.7} & 71.0\eb{0.5} & \underline{71.4\eb{0.7}} & \underline{71.8\eb{0.7}} \\
 & \single{} & 83.2\eb{0.6} & 83.2\eb{0.6} & 83.2\eb{0.6} & 83.2\eb{0.6} & 68.6\eb{0.6} & 68.6\eb{0.6} & 68.6\eb{0.6} & 68.6\eb{0.6} \\
 & Thicket & \underline{83.9\eb{0.0}} & \underline{84.4\eb{0.2}} & \textbf{84.9\eb{0.8}} & 84.2\eb{0.9} & \underline{71.2\eb{0.0}} & \underline{71.2\eb{0.5}} & 71.3\eb{0.7} & \textbf{71.9\eb{0.4}} \\
\addlinespace[2pt]
Llama-3.1-8B & Base & 69.6\eb{0.8} & 69.6\eb{0.8} & 69.6\eb{0.8} & 69.6\eb{0.8} & \textbf{66.2\eb{0.4}} & \underline{66.2\eb{0.4}} & \textbf{66.2\eb{0.4}} & \underline{66.2\eb{0.4}} \\
 & \singlem{} & \underline{74.0\eb{0.4}} & \textbf{76.6\eb{0.5}} & 72.2\eb{0.8} & 69.7\eb{0.8} & 64.2\eb{1.8} & 64.9\eb{0.4} & 65.6\eb{0.9} & 66.1\eb{0.1} \\
 & \single{} & 72.8\eb{1.4} & 72.8\eb{1.4} & \underline{72.8\eb{1.4}} & \underline{72.8\eb{1.4}} & 64.3\eb{1.4} & 64.3\eb{1.4} & 64.3\eb{1.4} & 64.3\eb{1.4} \\
 & Thicket & \textbf{74.6\eb{1.7}} & \underline{75.7\eb{1.5}} & \textbf{75.3\eb{0.9}} & \textbf{74.1\eb{0.6}} & \underline{65.2\eb{0.5}} & \textbf{67.2\eb{0.2}} & \underline{66.0\eb{0.1}} & \textbf{67.1\eb{0.3}} \\
\bottomrule
\end{tabular}

  \end{center}
\end{table}

\clearpage

\section{Disjoint random shards vs.\ subject-specialist shards}
\label{app:subject}

\textbf{Setup.} This ablation (Section~\ref{sec:subject}) asks whether the thicket's diversity should be \emph{engineered} rather than random. The MATH training pool is partitioned by its subject metadata into seven specialist shards (Algebra, Counting \& Probability, Geometry, Intermediate Algebra, Number Theory, Prealgebra, Precalculus), equalized at $184$ examples per adapter (the size of the smallest subject). The control thicket uses seven \emph{random} disjoint shards of the same size, drawn from the same $2{,}000$-problem pool. Both arms use the GRPO recipe of Table~\ref{tab:hyper}, so each member trains for $46$ steps. Each member contributes $23$ samples per problem ($7 \times 23 = 161$ completions per problem), and pools are scored with the protocol of Appendix~\ref{app:eval}, including confidence-filtered voting. The ablation uses a single seed ($42$) for four models.

\textbf{Results.} Table~\ref{tab:subject_perbench} breaks Table~\ref{tab:subject} down by benchmark. Across the $20$ (model, benchmark) cells, random shards give the higher majority vote in $11$, subject shards in $6$, and $3$ tie. Subject shards do best on Countdown for Qwen2.5-3B ($+2.2$ p.p.), while random shards do best on Knights~\&~Knaves ($+1.4$ to $+3.8$ p.p.\ on three models). The specialists' only consistent edge is a small one on GSM8K (subject $\geq$ random on all four models, by at most $0.3$ p.p.). On MATH500, despite its matching subject labels, random shards are never worse.

\begin{table}[!htbp]
  \caption{Per-benchmark majority vote and confidence-filtered vote (\%) for subject-specialist vs.\ random disjoint shards at $K{=}7$; single seed ($42$). Bold: better arm per row and vote type (ties bold both).}
  \label{tab:subject_perbench}
  \begin{center}
  \small
  \setlength{\tabcolsep}{5pt}
  \begin{tabular}{llcccc}
\toprule
 & & \multicolumn{2}{c}{Majority vote} & \multicolumn{2}{c}{Filtered vote} \\
\cmidrule(lr){3-4} \cmidrule(lr){5-6}
Model & Benchmark & Subject & Random & Subject & Random \\
\midrule
Qwen2.5-1.5B & MATH500 & \textbf{69.4} & \textbf{69.4} & 71.3 & \textbf{71.5} \\
 & GSM8K & \textbf{86.0} & 85.8 & 85.5 & \textbf{85.8} \\
 & OlympiadBench & 32.6 & \textbf{33.6} & 35.4 & \textbf{37.8} \\
 & Countdown & 49.3 & \textbf{49.8} & 49.3 & \textbf{49.8} \\
 & Knights \& Knaves & 33.8 & \textbf{35.2} & 34.4 & \textbf{35.2} \\
 & \emph{Mean} & 54.2 & \textbf{54.8} & 55.2 & \textbf{56.0} \\
\addlinespace[2pt]
Qwen2.5-3B & MATH500 & 77.4 & \textbf{77.5} & \textbf{78.7} & 78.2 \\
 & GSM8K & \textbf{91.4} & 91.1 & 91.2 & \textbf{91.4} \\
 & OlympiadBench & 43.3 & \textbf{44.2} & \textbf{44.8} & 44.2 \\
 & Countdown & \textbf{47.5} & 45.3 & \textbf{47.5} & 45.3 \\
 & Knights \& Knaves & 54.9 & \textbf{57.3} & \textbf{54.9} & 54.5 \\
 & \emph{Mean} & 62.9 & \textbf{63.1} & \textbf{63.4} & 62.7 \\
\addlinespace[2pt]
Qwen2.5-7B & MATH500 & 82.8 & \textbf{83.3} & \textbf{83.2} & \textbf{83.2} \\
 & GSM8K & \textbf{94.4} & \textbf{94.4} & 94.2 & \textbf{94.4} \\
 & OlympiadBench & 52.1 & \textbf{52.9} & \textbf{53.1} & 52.3 \\
 & Countdown & \textbf{54.6} & 54.2 & \textbf{54.6} & 54.2 \\
 & Knights \& Knaves & \textbf{61.7} & \textbf{61.7} & 61.7 & \textbf{63.7} \\
 & \emph{Mean} & 69.1 & \textbf{69.3} & 69.3 & \textbf{69.5} \\
\addlinespace[2pt]
Llama-3.1-8B & MATH500 & 66.5 & \textbf{66.7} & 67.0 & \textbf{68.5} \\
 & GSM8K & \textbf{91.7} & 91.4 & \textbf{91.5} & 91.2 \\
 & OlympiadBench & \textbf{35.3} & 35.2 & \textbf{36.4} & 36.0 \\
 & Countdown & 53.2 & \textbf{53.7} & 52.0 & \textbf{56.5} \\
 & Knights \& Knaves & 38.9 & \textbf{42.7} & \textbf{42.8} & 42.7 \\
 & \emph{Mean} & 57.1 & \textbf{57.9} & 57.9 & \textbf{59.0} \\
\bottomrule
\end{tabular}

  \end{center}
\end{table}

\end{document}